\documentclass[preprint,12pt]{elsarticle}

\usepackage{amssymb}
\usepackage[ruled,vlined]{algorithm2e}
\usepackage{amsmath}
\usepackage{multirow}
\usepackage{booktabs}
\usepackage{tabularx}
\usepackage{caption}
\usepackage{graphicx} 
\usepackage{makecell} 
\usepackage{array} 
\usepackage{caption}
\usepackage{float}
\usepackage{placeins}
\usepackage{subcaption}
\usepackage{graphicx}
\usepackage{url}
\journal{Applied Soft Computing}

\begin{document}

\begin{frontmatter}


\title{Attention-Driven Multimodal Alignment for Long-term Action Quality Assessment}

\author[label1]{Xin Wang} 
\ead{wwwwang5041359@gmail.com}
\author[label1]{Peng-Jie Li}
\ead{lipj@bsu.edu.cn}

\author[label1]{Yuan-Yuan Shen\corref{corresponding}}
\ead{shenyuanyuan@bsu.edu.cn}

\affiliation[label1]{organization={School of Sport Engineering, Beijing Sport University},
    city={Beijing},
    postcode={100084},
    country={China}}
\cortext[corresponding]{Corresponding author.}

\begin{abstract}
Long-term action quality assessment (AQA) focuses on evaluating the quality of human activities in videos lasting up to several minutes. This task plays an important role in the automated evaluation of artistic sports such as rhythmic gymnastics and figure skating, where both accurate motion execution and temporal synchronization with background music are essential for performance assessment. However, existing methods predominantly fall into two categories: unimodal approaches that rely solely on visual features, which are inadequate for modeling multimodal cues like music; and multimodal approaches that typically employ simple feature-level contrastive fusion, overlooking deep cross-modal collaboration and temporal dynamics. As a result, they struggle to capture complex interactions between modalities and fail to accurately track critical performance changes throughout extended sequences. To address these challenges, we propose the Long-term Multimodal Attention Consistency Network (LMAC-Net). LMAC-Net introduces a multimodal attention consistency mechanism that explicitly aligns features across different modalities, enabling stable integration of complementary multimodal information and significantly enhancing feature representation capabilities. Specifically, a multimodal local query encoder module with learnable queries is designed to automatically capture temporal semantics within each modality while dynamically modeling complementary relationships across modalities. To ensure interpretable evaluation results, we adopt a two-level score evaluation module, where stage-wise scores are first calculated to generate a final overall score. Additionally, we apply attention-based feature-level and regression-based result-level loss to jointly optimize multimodal alignment and decision-layer fusion. Experiments conducted on the RG and Fis-V datasets demonstrate that LMAC-Net significantly outperforms existing methods, validating the effectiveness of our proposed approach.
\end{abstract}


\end{frontmatter}

\section{Introduction}
AQA is a challenging task in fine-grained video understanding. It aims to automatically evaluate the execution quality of actions by identifying subtle differences in motion, and is widely applied in competitive sports for technical scoring and assisted training. Most existing approaches formulate AQA as a regression problem that maps video features to continuous quality scores.

Based on the duration of evaluation, AQA tasks are generally categorized into short-term AQA\cite{xu2022finediving,gedamu2023fine} and long-term AQA\cite{11,13}. Short-term AQA targets brief actions lasting a few to tens of seconds, such as diving, where performance evaluation typically relies on fixed-pattern segment analysis. In contrast, long-term AQA focuses on athletic performances with durations exceeding one minute, as seen in sports such as figure skating and floor gymnastics. These performances are composed of multiple interdependent technical segments, which often exhibit significant temporal variability and intricate dependencies. Effectively modeling these long-range temporal relationships is an important challenge for long-term AQA. 

To address this challenge, recent works have introduced transformer-based architectures with learnable queries in the decoder\cite{12,14,ji2023localization}, which enable the model to selectively attend to semantically important moments in long video sequences. These methods have shown strong performance in capturing global temporal structure and enhancing the accuracy of score prediction in long-term AQA tasks.

However, despite their effectiveness in temporal modeling, these approaches primarily rely on visual information alone. In many long-duration artistic sports, such as figure skating and rhythmic gymnastics, performance quality is not determined solely by movement execution, but also by the synchronization between movement and musical rhythm. This highlights an additional challenge in long-term AQA: the need to integrate multimodal information, particularly from audio and visual streams, to comprehensively assess performance. Although some recent studies have attempted to incorporate multiple modalities into AQA \cite{xia2023skating,20}, most perform feature fusion only at the global level, overlooking temporal variations in individual modalities. Consequently, these approaches cannot effectively verify the temporal alignment between visually salient actions (e.g., jumps/spins) and their corresponding acoustic signatures (e.g., musical accents or landing sounds), leading to suboptimal cross-modal understanding.

These limitations underscore the need for a more fine-grained, temporally aligned multimodal modeling framework. To this end, we propose LMAC-Net, a novel architecture that integrates a transformer-based multimodal local query encoder module and an attention-based feature-level loss. This design not only enables fine-grained analysis of dynamic changes in each modality over long temporal sequences, but also explicitly enforces the consistency of attention regions across different modalities at key moments. As a result, it enhances deep multimodal collaboration and feature alignment, thereby improving the model’s ability to understand and accurately score the complex interactions between movements and music in artistic sports performances.

Unlike previous multimodal approaches that rely on joint modeling of multiple modalities, which often causes interference between modalities and hinders the extraction of discriminative modality-specific features\cite{owens2018audio,wu2019dual,xiao2020audiovisual, 19,jiang2022egocentric}, our proposed encoder module adopts a modality-specific architecture. It consists of independent branches, each employing a transformer decoder with learnable queries to extract features from its corresponding modality. This design allows each branch to focus on the unique characteristics of its own modality and effectively preserve modality-specific information.

Moreover, interpretability is particularly critical in long-term AQA. In performance-oriented sports such as figure skating and gymnastics, providing only an overall score is often insufficient, as detailed feedback is essential for understanding performance quality at different stages. To enhance the interpretability of our proposed AQA model, we design a fine-to-coarse two-level score evaluation module after multimodal fusion. This module starts with detailed assessments within local temporal windows and progressively aggregates these into a higher-level evaluation over longer durations. Finally, a regression-based result-level loss is used to optimize the scoring performance of the model.

\begin{figure}[t]
\centering
\includegraphics[width=0.75\columnwidth]{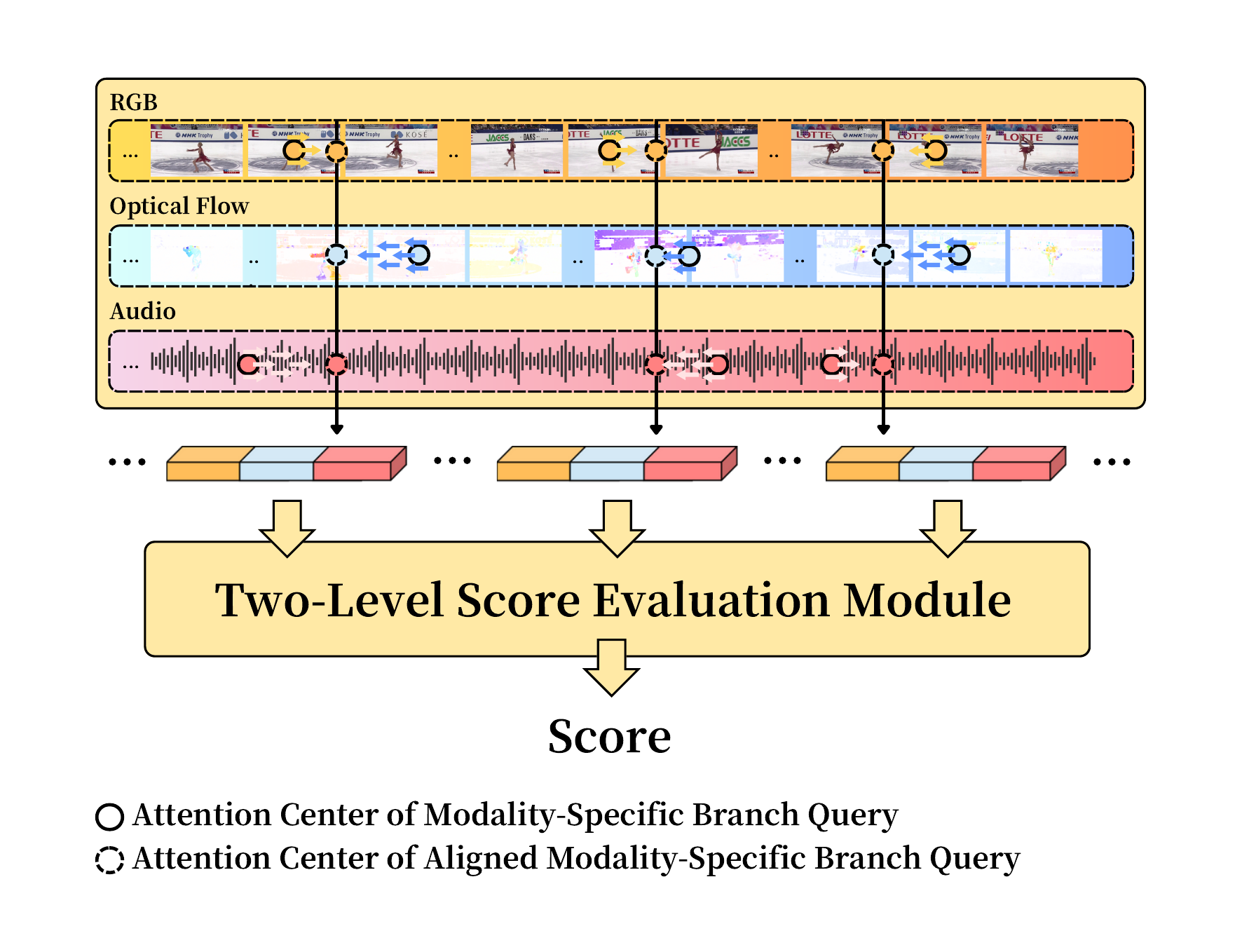}
    \caption{The core idea of the proposed LMAC-Net. In the multimodal local query encoder module, modality-specific branches with learnable queries are used to disentangle the key action segments that each modality attends to over time. Under the constraint of multimodal consistency loss, the alignment and integration of RGB, optical flow, and audio features are achieved by minimizing the distance between the attention centers of modality-specific queries. The final quality score is obtained through a multi-stage, fine-grained evaluation of the fused multimodal features.}
    \label{fig:Figure_1}
\end{figure}

We conducted experiments on two publicly available long-term video datasets: the rhythmic gymnastics RG dataset\cite{13} and the figure skating Fis-V dataset\cite{11}. Experimental results show that the proposed LMAC-Net significantly outperforms existing mainstream methods. Extensive ablation studies further verify the effectiveness of the designed modules and fully showcase the advantages of our method in multimodal learning. The main idea of our approach is illustrated in Figure \ref{fig:Figure_1}.

The contributions of our work are threefold:
\begin{itemize}
\item We propose a multimodal local query encoder module for feature extraction, which performs temporal parsing across modality branches to capture dynamic changes and identify key features for long-term AQA.
\item We design a fine-to-coarse two-level score evaluation module that independently regresses scores for local temporal windows and computes the final score through adaptive weighting, which enhances the model’s interpretability.
\item We propose a method that explicitly captures cross-modal temporal consistency by combining temporal parsing with attention-based local alignment. This design improves the coordination of multimodal features in critical action stages, leading to more accurate AQA performance. 
\end{itemize}

The rest of the paper is organized as follows. In Section 2, we briefly discuss some related work. Section 3 provides our approach in detail. We describe the datasets and present experimental results in Section 4. Finally, Section 5 concludes this paper.

\section{Related Work}
Our work is closely related to three major subfileds of machine learning research: unimodal action quality assessment, multimodal action quality assessment and attention-based cross-modal temporal alignment. Below we briefly review representative work in each subfield. A comparative summary of representative long-term AQA methods and their technical characteristics in provided in  Table~\ref{tab:aqa_review} for reference.

\subsection{Unimodal Action Quality Assessment}
Unimodal Action Quality Assessment (AQA) is formulated as a fine-grained score regression task and was first introduced by Pirsiavash et al. \cite{1}, who employed handcrafted skeleton features for performance scoring. However, this approach showed limited generalization in complex scenarios. With the rise of deep learning, the field has shifted from manually designed motion descriptors to automatic modeling of visual features using convolutional and recurrent networks. Existing unimodal AQA methods can be broadly classified into short-term and long-term methods based on video duration.

Short-term AQA methods mainly focus on brief actions such as diving. Parmar et al. \cite{2} proposed a model based on C3D and LSTM that significantly improved evaluation accuracy. Li et al. \cite{3} introduced the Scoring Net, combining 3D convolution and bidirectional LSTM to achieve action semantic segmentation and phased evaluation. Xiang et al. \cite{4} further enhanced accuracy by leveraging the S3D segmentation network to capture local details of actions. Recently, Parmar et al. \cite{5} introduced a multi-task framework, Jain et al. \cite{6} incorporated contrastive learning through a Siamese network, Yu et al. \cite{7} transformed score regression into a classification and regression task, and Wang et al. \cite{8} proposed the TSA module, which aggregates features using a pipeline mechanism and self-attention. These methods have collectively strengthened model interpretability and robustness.

Long-term AQA methods typically focus on prolonged action sequences such as figure skating and rhythmic gymnastics. Xu et al. \cite{11} introduced a self-attention LSTM framework to effectively capture both local and global features. In contrast, Doughty et al. \cite{49} proposed a ranking-aware temporal attention model to isolate high-skill and low-skill segments, addressing the issue of information loss in long videos. More recently, Zeng et al. \cite{13} presented Action-Net, which combines graph neural networks with local context information to generate global features. Building on similar idea, Zhang et al. \cite{zhang2023logo} proposed the GOAT method, which integrates attention mechanisms and graph convolutional networks to jointly model segmented long-term temporal features and intra-group relationships, thereby enabling efficient capture of complex temporal actions. Additionally, Xu et al. \cite{14} proposed GDLT, which uses query mechanisms to capture complex dependencies. Ji et al. \cite{ji2023localization} disentangled the features of PCS and TES through sub-action localization and uncertainty modeling, enabling efficient and accurate modeling for long-term AQA tasks.

While unimodal AQA methods have achieved promising results, they are inherently limited by their reliance on a single modality, which restricts their ability to capture the full semantic and contextual richness of complex athletic performances. In particular, long-duration tasks often involve factors such as rhythm, context, and external cues that are difficult to infer from visual or motion features alone. These limitations have motivated a growing body of research on multimodal AQA, which seeks to incorporate complementary sources of information to enhance evaluation accuracy and interpretability.

\subsection{Multimodal Action Quality Assessment}
In response to these challenges, researchers have increasingly turned to multimodal AQA methods, which integrate additional data sources beyond visual appearance or motion. Building upon recent advances in multimodal video understanding \cite{45,46,47,48}, these approaches incorporate signals such as audio, text, and inertial measurements to improve robustness and semantic richness. A number of representative works have demonstrated the effectiveness of multimodal fusion in capturing complex patterns in athletic performances.

One line of research focuses on incorporating auxiliary sensor data. For example, to handle complex outdoor scenarios such as windsurfing, Nagai et al.\cite{30} employed multi-head attention mechanisms within a transformer framework to fuse visual features with auxiliary modalities like IMU and GPS, significantly enhancing the robustness of the model in complex actions and multi-view conditions.

Another major direction involves integrating textual information to provide high-level semantic guidance. Xu et al.\cite{xu2024vision} proposed a multi-granular visual-semantic alignment framework based on CLIP text semantics and semantic-aware collaborative attention, enabling deep interaction between visual and semantic features at the video, stage, and frame levels, and thereby achieving more precise semantic understanding of diving actions. Du et al. \cite{du2023learning} introduced the semantic-guided network (SGN), which uses a teacher-student framework to jointly encode figure skating commentary text and video features into semantic-aware representations. By employing learnable atomic queries and multiple alignment losses, semantic knowledge is efficiently transferred to the visual branch, enabling high-precision figure skating video assessment using only visual input.

In long-duration artistic sports, audio information also plays a crucial role, as movement rhythms often align closely with musical beats. Addressing this, Xia et al. \cite{xia2023skating} proposed the first audio-visual multimodal framework, Skating-Mixer, which learns joint audio and visual representations using memory recurrent units and multilayer MLP modules, enabling effective fusion and scoring of long figure skating videos. However, this direct joint modeling approach can lead to interference between modalities and limit the expressiveness of multimodal features. To address this limitation, Zeng et al.\cite{20} designed the Progressive Adaptive Multimodal Fusion Network (PAMFN), which utilizes multi-stage feature decoding and adaptive fusion strategies to separately model single-modality and mixed-modality features, thereby effectively improving the accuracy of figure skating and gymnastics scoring.

Despite these advances, most existing multimodal AQA methods rely on aligning global feature representations across modalities, often neglecting the temporal consistency of local attention regions. This limits the ability of individual modality branches to collaboratively focus on key action segments, making it difficult to capture fine-grained dynamic interactions, such as the synchronization of movement and music over long temporal sequences. For long-term AQA, accurately modeling cross-modal temporal relationships remains a key challenge.

\setlength{\tabcolsep}{4pt}
\renewcommand{\arraystretch}{1.25}
\begin{table}[!t]
\footnotesize
\caption{Representative methods for long-term AQA and their characteristics.}
\label{tab:aqa_review}
\resizebox{\textwidth}{!}{
\begin{tabular}{|c|c|c|c|c|c|c|}
\hline
\textbf{Method} & \textbf{Modality} & \textbf{Temporal Modeling} & \textbf{Alignment/Fusion} & \textbf{Innovation/Remark} \\
\hline
Xu et al. (C3D+SVR )~\cite{2} & RGB & C3D+LSTM & - & Local and global temporal modeling \\
\hline
Zeng et al. (Action-Net)~\cite{4} & RGB & GCN+Attention & - & Context-aware, GNN \\
\hline
Xu et al. (GDLT)~\cite{5} & RGB & Query-based Transformer &  -  & Action-grade feature disentanglement, Fine-grained score modeling \\
\hline
Ji et al. (LUSD-Net)~\cite{6} & RGB & Query-based Transformer+Sub-action localization &  -  & PCS/TES feature disentanglement \\
\hline
Du et al. (SGN)~\cite{du2023learning} & RGB, Text & Query-based Transformer &  Cross-modal Attention Fusion & Commentary-guided semantic distillation \\
\hline
Xia et al. (Skating-Mixer)~\cite{7} & RGB, Audio & RNN+MLP & Joint Fusion & MLP-based, Audio-visual joint temporal modeling \\
\hline
Zeng et al. (PAMFN)~\cite{8} & RGB, Flow, Audio & Multi-stage ConvBlock & Modal-specific Progressive Fusion & Multi-branch,  Staged adaptive fusion \\
\hline
 Ours (LMAC-Net) &  RGB, Flow, Audio &  Query-based Transformer+Temporal Parsing  &  Modal-specific Temporal Collaborative Alignment &  Multimodal temporal parsing, Fine-grained cross-modal feature alignment,  Interpretable scoring \\
\hline
\end{tabular}
}
\end{table} 

\subsection{Attention-Based Cross-Modal Temporal Alignment}
A major challenge in multimodal video understanding lies not only in aligning the semantic features of heterogeneous modalities within a unified representation space, but also in achieving accurate cross-modal temporal alignment. Temporal synchronization encodes rich structural dependencies that are essential for interpreting complex events. For instance, the temporal correspondence between audio and visual streams not only reflects event dynamics but also provides valuable supervisory signals for guiding feature alignment and fusion. Leveraging such temporal structure facilitates effective cross-modal consistency learning \cite{mercea2022temporal} and enhances multimodal integration. While global semantic alignment in embedding space improves inter-modal consistency, it is insufficient to ensure temporal synchronization across modalities for the same event. This highlights the need for models that explicitly capture and utilize cross-modal temporal relationships to enable fine-grained and coherent understanding of dynamic scenes.

To address this, an increasing number of methods have introduced attention-based cross-modal temporal alignment. Generally, these approaches employ cross-attention or its variants to dynamically associate sequences across visual, audio, or textual modalities. For example, in audio-visual alignment tasks, Mercea et al.\cite{mercea2022temporal} proposed a repeated cross-modal attention module that only allows visual and audio features to attend to the corresponding sequence in the other modality, thereby enhancing their temporal correlation. Praveen et al.\cite{praveen2024cross} introduced the dynamic cross-attention mechanism, which adaptively controls the strength of cross-modal fusion over time, improving the robustness and consistency of multimodal fusion.

Further, some methods build upon attention mechanisms by introducing explicit alignment modules or alignment losses to reinforce temporal synchronization. Recently, in audio-visual speech recognition, Liu et al.\cite{liu2024alignvsr} proposed AlignVSR, which achieves fine-grained correspondence between visual frames and audio units through a combination of global cross-modal alignment and frame-level local alignment mechanisms. Similarly, Hu et al.\cite{hu2023cross} introduced GILA, which models global cross-modal complementary relationships via attention and uses contrastive learning losses for frame-level temporal consistency between audio and visual features, thereby enhancing multimodal representation. These approaches aim to capture deep complementary relationships and temporal consistency across modalities.

In the area of video-text alignment, Han et al.\cite{han2022temporal} proposed the temporal alignment network, designing a multimodal transformer joint encoder combined with complementary dual-encoder co-training to achieve frame-level temporal alignment between long videos and textual descriptions. To address fine-grained alignment between long videos and text, LF-VILA\cite{sun2022long} introduced a temporal contrastive loss that aligns video–text similarity with their temporal proximity. This loss explicitly encodes temporal order to improve cross-modal consistency.

While attention-based cross-modal alignment methods have advanced multimodal video understanding, their application to AQA remains limited. Most existing multimodal AQA approaches still focus on global feature alignment or simple contrastive learning, lacking the fine-grained temporal synchronization needed for accurate long-term action evaluation. Addressing these challenges requires more precise modeling of stage-wise and segment-level temporal consistency across modalities.

\section{Methodology}
In this section, we first provide a general introduction to LMAC-Net, our proposed multimodal AQA framework (Algorithm~\ref{alg:lmacnet}). Next, in Section 3.1, we delve into the two key components of our approach: the multimodal local query encoder module (Section 3.2) and the two-level score evaluation module (Section 3.3). Finally, we present and discuss the objective function in Section 3.4.

\begin{algorithm}[t]
\SetAlgoNoLine
\caption{Framework of LMAC-Net}
\label{alg:lmacnet}
\KwIn{RGB frames, optical flow, audio signals}
\KwOut{Action quality score}
\BlankLine
1.Divide the RGB frames, optical flow, and audio into the same number of non-overlapping segments.\\
2.Extract segment-level features for each modality using pre-trained backbones:\\
\Indp
\textbullet~VST for RGB, I3D for optical flow, AST for audio.\\
\Indm
3.For each modality-specific feature: \\
\Indp
\textbullet~Apply temporal parsing to model long-term temporal structure;\\
\textbullet~Enforce multimodal consistency constraints for cross-modal alignment. \\
\Indm
4.Concatenate the modality-specific query features from all three modalities for each query. \\
5.Feed the fused features into the two-level score evaluation module to predict the final action quality score: \\
\Indp
\textbullet~First, regress intermediate query-level scores;\\
\textbullet~Then, fuse the query-level scores into a global action quality score.\\
\end{algorithm}

\subsection{Overview}
AQA is essentially a regression problem, analogous to the scoring process used by judges in sports competitions. The goal is to predict a non-negative real-valued score that reflects the quality of an action. The core challenge of AQA lies in designing an effective model that accurately learns the mapping between action data and their corresponding scores.

In this paper, we address the AQA task in a multimodal context. Specifically, LMAC-Net uses both video and audio inputs, including RGB frames, optical flow, and audio signals, to capture complementary visual and auditory features. The model is trained to map these multimodal inputs to quality scores under the supervision of expert annotations.

To process the input data, we divide the RGB frames and optical flow modalities into non-overlapping segments of equal length, with each segment comprising several consecutive frames. Similarly, the audio data is segmented into the same number of segments to ensure alignment with the RGB and optical flow modalities. Following \cite{20}, RGB features are extracted using VST\cite{31}, optical flow features are extracted using I3D\cite{32}, and audio features are extracted using  AST\cite{34}. These backbone models were pre-trained on the Kinetics-600\cite{32}, Kinetics-400\cite{32}, and AudioSet datasets\cite{33}, respectively.

Formally, the feature sequences for the three modalities are defined as follows:
\begin{equation}
\begin{aligned}
    F_{RGB} &= \{f_t^{RGB}|f_t^{RGB}\in \mathbb{R}^{d_{RGB}},t=1,2,\dots,T\}, \\
    F_{Flow} &= \{f_t^{Flow}|f_t^{Flow}\in \mathbb{R}^{d_{Flow}},t=1,2,\dots,T\}, \\
    F_{Audio} &= \{f_t^{Audio}|f_t^{Audio}\in \mathbb{R}^{d_{Audio}},t=1,2,\dots,T\},
\end{aligned}
\end{equation}
where $T$ is the temporal length of the sequences, and $d_{RGB}$, $d_{Flow}$, $d_{Audio}$ are the feature dimensions for the RGB, optical flow, and audio modalities, respectively.

These sequences are processed by the multimodal local query encoder module with modality-specific learnable queries, denoted as $\varepsilon_{query}(\cdot)$. This encoder captures temporal semantics and explores cross-modal complementarity, transforming the input features into a unified representation space. The output of the encoder is the multimodal feature $F_{multi}$, expressed as:
\begin{equation}
    F_{multi} = \varepsilon_{query}(F_{RGB}, F_{Flow}, F_{Audio}).
\end{equation}

The multimodal feature $F_{multi}$ is then input into the two-level score evaluation module, denoted as $\psi_{score}(\cdot)$, to predict the action quality score $S\in\mathbb{R}$ as follows:
\begin{equation}
    S = \psi_{score}(F_{multi}).
\end{equation}

\begin{figure}[t]
    \centering
\includegraphics[width=1\columnwidth]{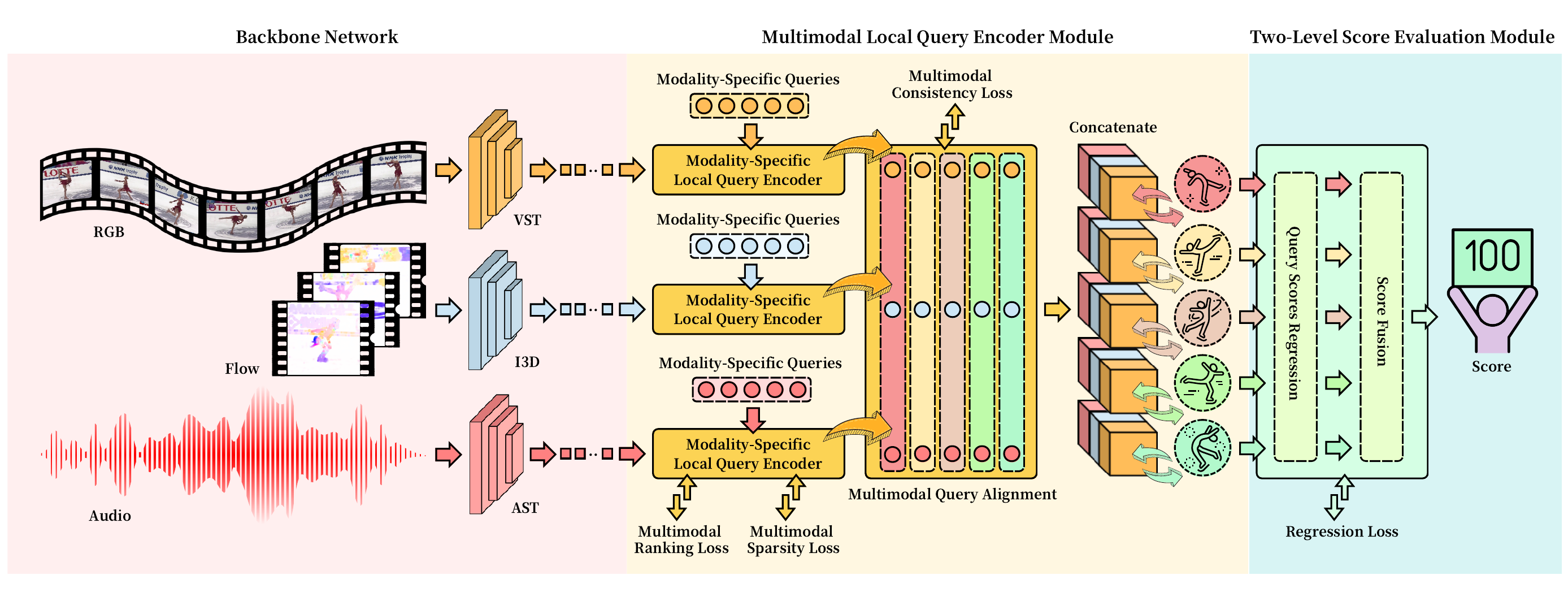}
    \caption{The proposed framework of LMAC-Net. RGB, optical flow, and audio input are processed by pre-trained backbone networks for feature extraction. A multimodal local query encoder captures temporal semantics and explores cross-modal complementarity. Finally, a two-level evaluation module performs fine-grained score regression on the fused multimodal features.}
    \label{fig:Figure_2}
\end{figure}

The detailed descriptions of $\varepsilon_{query}(\cdot)$, $\psi_{score}(\cdot)$ and the loss functions are provided in
Section 3.2, Section 3.3 and Section 3.4. A summary of the entire process is illustrated in Figure \ref{fig:Figure_2}.

\begin{figure}[t]
    \centering
\includegraphics[width=0.55\columnwidth]{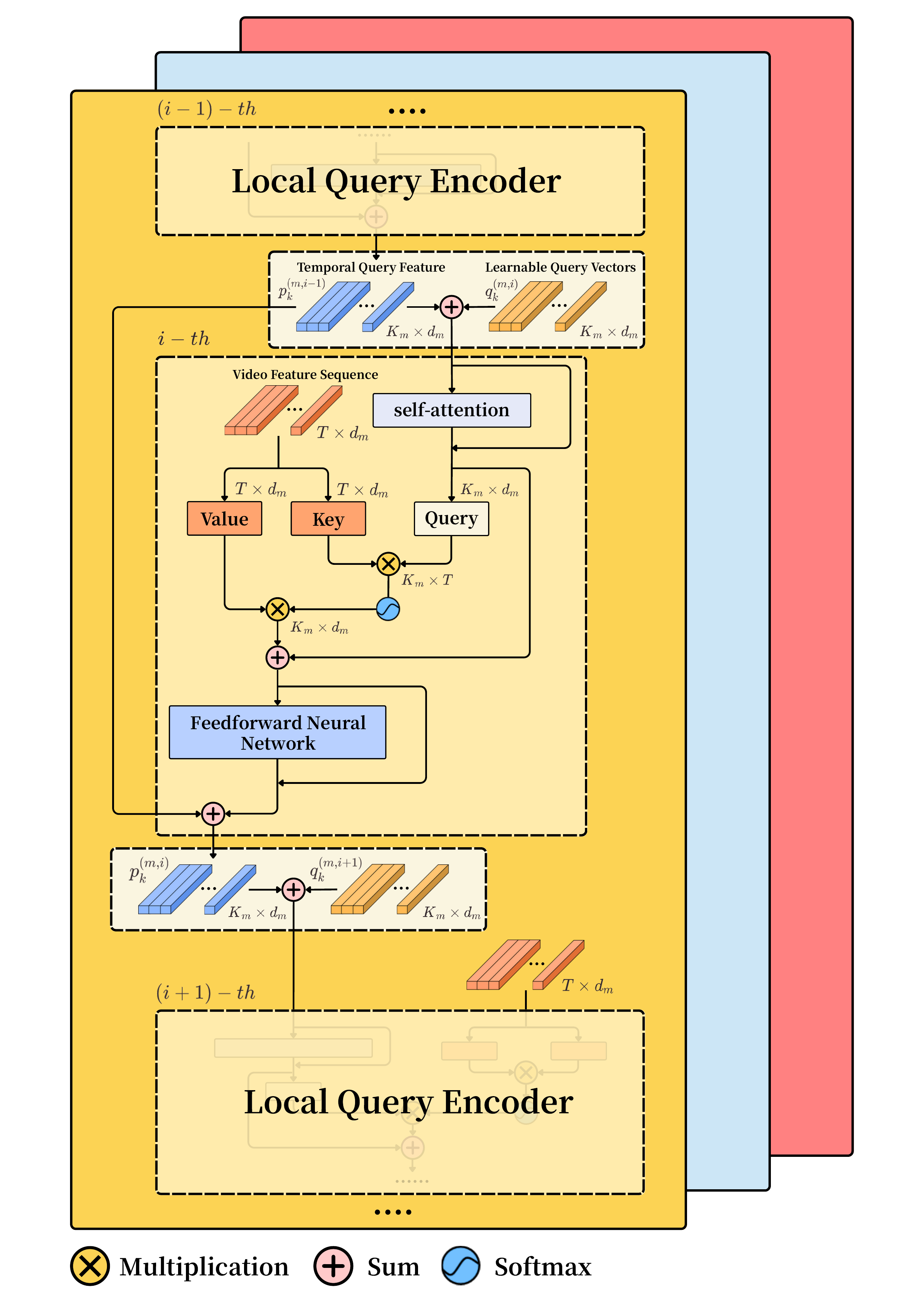}
\caption{Illustration of our multimodal local query encoder module. The module contains three independent branches for RGB, optical flow, and audio modalities, each consisting of multiple decoder layers. Important tensor dimensions are annotated in the figure for better understanding. $Query$, $Key$, and $Value$ represent three different linear projections. }
\label{fig:Figure_3}
\end{figure}

\subsection{Multimodal Local Query Encoder Module}
The multimodal local query encoder module consists of two key components: three independent modality-specific branches and a feature fusion process. Each branch extracts temporal features for a specific modality using a shared architecture with learnable query vectors and self-attention mechanisms to capture fine-grained temporal patterns. These features are then fused to form a unified representation that takes advantage of complementary cross-modal information.

For the three modality-specific branches, we adopt the same multi-layer transformer decoder architecture as \cite{9} to extract and analyze modality-specific features. Specifically, learnable query vectors are first utilized in a cross-attention mechanism to dynamically extract task-relevant semantic features from each modality. These features are then refined through self-attention, which captures contextual dependencies within each modality. The resulting temporal query features effectively represent key actions in long-term sports performances.

Figure \ref{fig:Figure_3} illustrates the feature extraction process within a single modality, detailing the structure of the modality- specific branches. Each branch corresponds to a specific modality $m\in\{RGB,Flow,Audio\}$ and consists of multiple stacked transformer decoders. In the $i$-th decoder layer for the $m$-th modality, a set of $K$ learnable atomic patterns, $\{q_k^{(m,i)}\}$, are combined with the temporal query features from the previous layer, $\{p_k^{(m,i-1)}\}$, to generate the query representations, $\{\widehat{q}_k^{(m,i)}\}$, where $k\in\{1,\cdots,K\}$. Specifically, the query representations are computed as:
\begin{equation}
\begin{aligned}
\widehat{q}_k^{(m,i)}={p}_k^{(m,i-1)}+{q}_k^{(m,i)},
k&\in \{1,\cdots,K\},
\\
m&\in\{RGB,Flow,Audio\}
\end{aligned}
\end{equation}
In the first layer, the temporal query features are initialized as zero vectors.

The query vectors are then used to perform cross-attention with input features, capturing local temporal dependencies. Queries, keys, and values are computed via linear projections as follows:
\begin{equation}
\begin{aligned}  \tilde{q}_k^{(m,i)}&=W_q^{(m,i)}\hat{q}_k^{(m,i)}, \\
k_t^{(m,i)}&=W_k^{(m,i)} f_t^{(m,i)}, \\
v_t^{(m,i)}&=W_v^{(m,i)} f_t^{(m,i)},
\end{aligned}
\end{equation}
where $W_q^{(m,i)}$, $W_k^{(m,i)}$, $W_v^{(m,i)}$$\in\mathbb{R}^{d_m\times d_m}$ are projection matrices, $d_m$ denotes the input feature dimension for modality $m$, and $f_t^{(m,i)}$ represents the input feature of the $t$-th temporal segment. 

During decoding, the attention weight $\alpha_{k,t}^{(m,i)}$ measures the attention assigned by query $k$ to temporal segment $t$:
\begin{equation}    \alpha_{k,t}^{(m,i)}=\frac{\exp{((\tilde{q}_k^{(m,i)})^Tv_t^{(m,i)}/\tau)}}{\sum_{j=1}^T\exp{((\tilde{q}_k^{(m,i)})^Tv_j^{(m,i)}/\tau)}}
\end{equation}
where $\tau$ is a learnable temperature parameter controlling the scaling of the inner product, ensuring distinctive attention values. The temporal query features are updated as:
\begin{equation}
p_k^{(m,i)}=\sum_{j=1}^Ta_{k,j}^{(m,i)}v_j^{(m,i)}+p_k^{(m,i-1)}
\end{equation}
To enhance feature expressiveness, a feed-forward neural network (FFN) is applied to the temporal query features $p_{k}^{(m,i)}$, enabling the capture of complex relationships. Additionally, a multi-head self-attention layer is further used to model interactions among temporal query features, improving the understanding of key actions in long-term actions.

Finally, to integrate temporal information across modalities, we concatenate the temporal query features from RGB, optical flow, and audio into the final multimodal temporal query feature ${p}_k$:
\begin{equation}
{p}_k=\begin{bmatrix}{p}_k^{{RGB}},{p}_k^{{Flow}},{p}_k^{{Audio}}\end{bmatrix}.
\end{equation}
This concatenation enhances the model's understanding of multimodal information in long-term videos.

\subsection{Two-Level Score Evaluation Module}
\begin{figure}[t]
    \centering    \includegraphics[width=0.7\columnwidth]{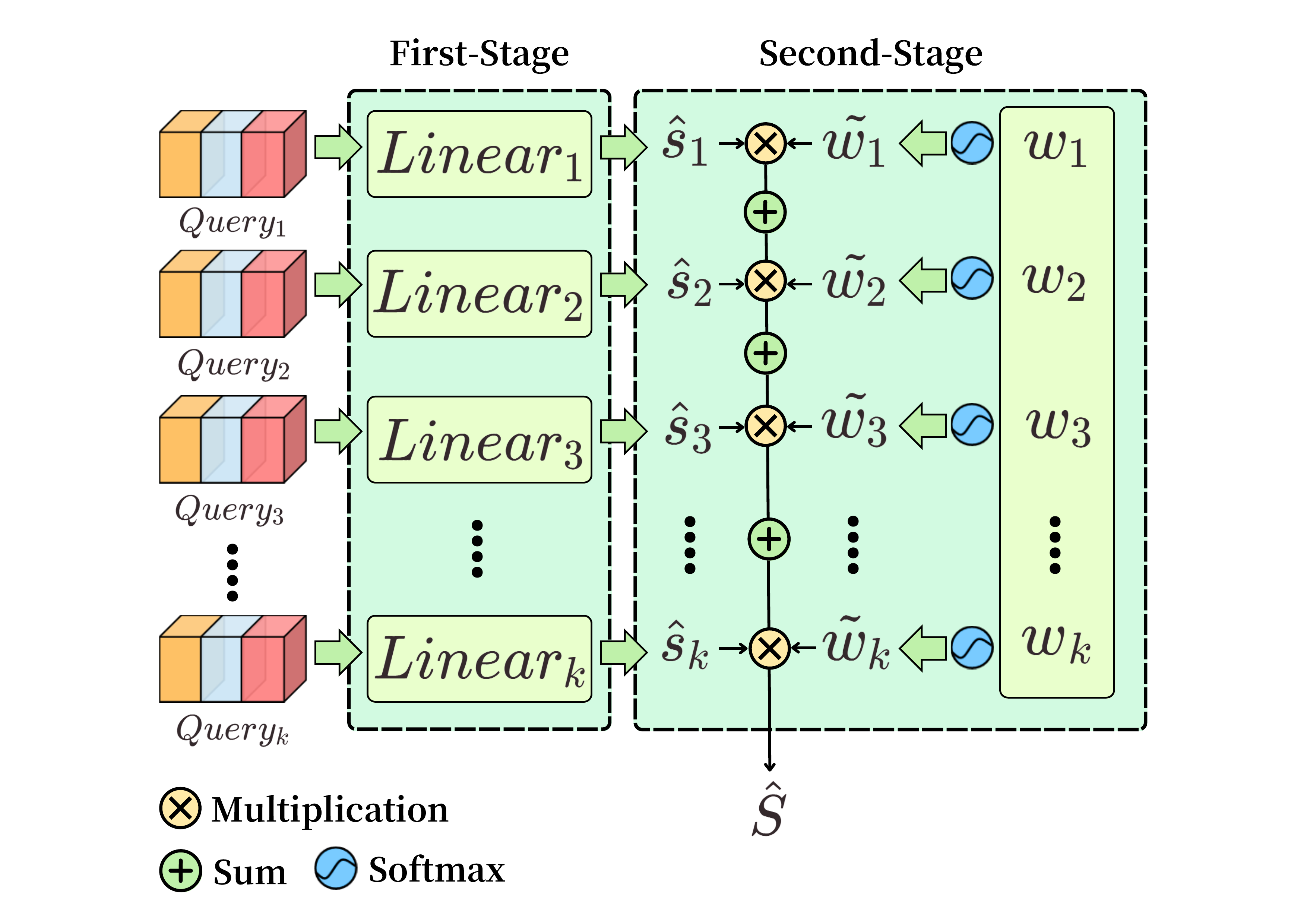}
\caption{Illustration of our two-level score evaluation module. \textbf{The first-stage evaluation} is based on the regression score of multimodal features for each query. \textbf{The second-stage evaluation} adaptively fuses the previous scores to obtain the final score. }
\label{fig:Figure_4}
\end{figure}

To achieve more granular assessment and improve interpretability, we propose a two-level score evaluation module for final action quality scoring based on the extracted temporal query features. In the first stage, we perform score regression on each multimodal temporal query feature to generate an initial score for each key temporal segment. In the second stage, a fusion strategy combines these initial scores to produce the final action quality score. The overall framework of our two-level evaluation module is shown in Figure \ref{fig:Figure_4}.

Formally, in the first stage, for each temporal query feature $p_{k}$, we use a linear regression layer to predict the score $\hat{s}_{k}$, which represents the evaluation result of the $k$-th temporal query feature. After this step, we obtain $K$ scores, denoted as $\{\hat{s}_1,\hat{s}_2,\ldots,\hat{s}_K\}$, which correspond to the model's assessment of different temporal segments.

In the second stage, we use another linear layer to aggregate these scores and produce the final overall action quality score. In this process, we introduce a learnable weight vector $\boldsymbol{w}\in\mathbb{R}^K$, which is normalized using the softmax function to ensure the sum of all weights equals 1:
\begin{equation}
    \tilde{\boldsymbol{w}} = \mathrm{softmax}(\boldsymbol{w})
\end{equation}
The final action quality score $\hat{S}$ is computed as:
\begin{equation}
   \hat{S}=\sum_{k=1}^K\tilde{w}_k\cdot \hat{s}_k
\end{equation}
where $w_k$ represents the normalized weight for the $k$-th score, and $\hat{s}_k$ is the score for the $k$-th temporal query feature.

Thus, our two-level score evaluation module allows the model to dynamically adjust the contribution of each temporal segment to the final score based on its importance, effectively focusing on the most critical segments for assessing action quality.

\subsection{Loss Functions}
To ensure effective feature learning and accurate decision-making, we adopt a composite loss function that operates on both the feature and decision levels. This dual-level design enhances the model's capacity to learn robust feature representations while directly optimizing the final task objective. The total loss function $\mathcal{L}$ is defined as follows:
\begin{equation}
    \mathcal{L} = \lambda_1 \mathcal{L}_{\text{score}} + \lambda_2 \mathcal{L}_{\text{feature}},
\end{equation}
where $\mathcal{L}_{\text{score}}$ represents the loss related to the task-specific output (i.e., action quality score), $\mathcal{L}_{\text{feature}}$ represents the loss related to learning robust feature representations, and $\lambda_1$ and $\lambda_2$ are hyperparameters that control the relative importance of the score and feature losses, respectively.

\textbf{Decision-Level Loss ($\mathcal{L}_{\text{score}}$)}
The decision-level loss $\mathcal{L}_{\text{score}}$ minimizes the discrepancy between the predicted action quality scores $\hat{S}_n$ and the ground truth scores $S_n$, ensuring accurate performance in task prediction. It is defined as a mean squared error (MSE) loss:
\begin{equation}
    \mathcal{L}_{\text{score}} = \frac{1}{N} \sum_{n=1}^N (\hat{S}_n - S_n)^2,
\end{equation}
where $N$ is the total number of samples, $\hat S_n$ and $S_n$ are the predicted action quality score and the groundtruth action quality score for sample $n$, respectively.

\textbf{Feature-Level Loss d($\mathcal{L}_{\text{feature}}$)} To enhance the robustness of individual modalities within their feature space and promote efficient collaboration between multimodal features, we introduce the feature-level loss $\mathcal{L}_{\text{feature}}$. This loss comprises three key components: the multimodal ranking loss $\mathcal{L}_{\text{rank}}$, the multimodal sparsity loss $\mathcal{L}_{\text{sparsity}}$, and the multimodal consistency loss $\mathcal{L}_{\text{consistency}}$. These components jointly address both intra-modal temporal parsing and inter-modal alignment, ensuring robust feature representation and efficient multimodal integration.

The final feature-level loss is expressed as:
\begin{equation}
\mathcal{L}_{\text{feature}} = \mathcal{L}_{\text{rank}} + \mathcal{L}_{\text{sparsity}} + \mathcal{L}_{\text{consistency}}.
\end{equation}


At the heart of these losses is an attention center mechanism \cite{9} that focuses on the most relevant temporal segments of the multimodal features. This attention mechanism computes the attention center for each query, which reflects the model's focus on different time segments. The attention center is defined as the weighted average of the attention weights assigned to temporal segments, capturing the model's focus across time.

For each query $k$ in modality $m$, the attention weights from all decoder layers are concatenated into a single vector for each time step. This results in the attention weight vector for query $k$ at time step $t$ as ${\alpha}_{k,t}^m$. The attention center $\bar{\alpha}_{k}^{m}$ for the $k$-th query in modality $m$ is calculated by taking the weighted average of the attention weights across all time steps:
\begin{equation}
\begin{aligned}
    &\bar{\alpha}_k^m = \sum_{t=1}^T t \cdot \alpha_{k,t}^m, \\
    s.t. &\sum_{t=1}^T\alpha_{k,t}^m=1
\end{aligned}
\end{equation}
Here, the attention weights $\alpha_{k,t}^m=1$ are normalized across all time steps.

Based on attention centers, we introduce the multimodal ranking loss, which ensures that the temporal order of key segments is maintained within each modality. This loss function enforces consistency in the temporal ordering of attention centers by penalizing any violation of the correct temporal order between adjacent segments. The loss is defined as:
\begin{align}
    L_{\mathrm{rank}} = \sum_{m=1}^M \lambda_{\mathrm{rank}}^m \bigg( & \sum_{k=1}^{K-1} \max(0, \bar{\alpha}_k^m - \bar{\alpha}_{k+1}^m + d) \nonumber \\
    & + \max(0, 1 - \bar{\alpha}_1^m + d) \nonumber \\
    & + \max(0, \bar{\alpha}_K^m - T + d) \bigg)
\end{align}
where $d$ is a hyperparameter that controls the boundary, and $\lambda_{\mathrm{rank}}^{m}$ is a weight factor for balancing the contribution of each modality to the ranking loss.

Additionally, the multimodal sparsity loss is introduced to ensure that each query's attention distribution is more concentrate on the most important temporal segments within each modalities. This loss aims to improve the model's ability to distinguish important features in long-term actions. It is defined as:
\begin{equation}
    L_{\mathrm{sparsity}} = \sum_{m=1}^M \lambda_{\mathrm{sparsity}}^m \left( \sum_{k=1}^K \sum_{t=1}^T |t - \bar{\alpha}_k^m| \cdot \alpha_{k,t}^m \right),
\end{equation}
where $\lambda_{\mathrm{sparsity}}^m$ is a weight factor for modality $m$, controlling the contribution of each modality to the sparsity loss.

One of the core ideas of LMAC-Net is to minimize the "distance" between the attention centers of different modalities, thereby measuring and enhancing the consistency of each modality branch’s focus on key action segments. This distance reflects the degree of similarity in the temporal positions where different modality branches concentrate their attention. A smaller distance indicates that different modalities are more likely to focus on the same or similar key action segments along the temporal axis, which facilitates stronger multimodal collaboration and feature alignment. In contrast, a larger distance means that the modalities are attending to more dispersed temporal segments, which weakens feature fusion and information complementarity. To this end, we introduce a multimodal consistency loss to ensure alignment between the attention centers of different modalities at the same temporal segments. This loss promotes consistency across modalities, encouraging them to focus on the same key segments of the action sequence. As a result, multimodal information is integrated more effectively. The loss is defined as:
\begin{equation}
    L_{\text{consistency}} = \sum_{t=1}^T\sum_{1 \leq i<j \leq M} \left( \left\| \bar{\alpha}_t^{m_i} - \bar{\alpha}_t^{m_j} \right\|^2 \right),
\end{equation}
where $\bar\alpha_t^{m_i}$ and $\bar\alpha_t^{m_j}$ represent the attention centers for modalities $m_i$ and $m_j$ at time step $t$, respectively.

\section{Experiments}
We evaluated our method on two publicly available benchmark datasets for long-term action quality assessment: Rhythmic Gymnastics (RG) dataset\footnote{\url{https://github.com/qinghuannn/ACTION-NET}} and Fis-V dataset\footnote{\url{https://github.com/chmxu/MS\_LSTM}}. First, in Section 4.1, we briefly introduce the datasets. Then, in Section 4.2, we detail the evaluation metrics used in the experiments. Section 4.3 describe the implementation details of the experiments, followed by the presentation of experimental results in Section 4.4. In Section 4.5, we explore the impact of different modules and loss functions on overall performance through ablation studies. Finally, in Section 4.6, we visualize the qualitative results to fully demonstrate the effectiveness of our method.
\subsection{Datasets}
 In this section, we provide detailed descriptions of the two benchmark datasets used in our experiments: the RG dataset and the Fis-V dataset. Both are widely adopted in the literature as standard benchmarks for long-term action quality assessment (AQA) and offer well-annotated samples with multimodal information. Their official descriptions and download instructions can be found in the corresponding publications \cite{13,11}.

\noindent\textbf{RG Dataset.} The RG dataset contains 1,000 videos of rhythmic gymnastics, covering four types: ball, clubs, hoop, and ribbon, with 250 videos for each type. The average duration of each video is about 1 minute and 35 seconds, with 25 fps. The videos are sourced from high-standard international rhythmic gymnastics competitions and are finely annotated, including difficulty score, execution score, and final score. Following the official guideline\cite{13}, we selected 200 videos from each type for training and 50 videos for testing. 

\noindent\textbf{Fis-V dataset.} The Fis-V dataset is a large-scale video dataset specifically designed for the scoring task of figure skating competitions. It contains 500 short program videos of female skaters from multiple high-standard international skating events. The average duration of each video is about 2 minutes and 50 seconds, with 25 fps. The dataset includes two scores given by nine international judges: the Total Element Score (TES) and the Program Component Score (PCS). We follow the official recommendation\cite{11} for splitting the dataset, with 400 videos for training and 100 videos for testing.

\subsection{Evaluation Metrics}
\noindent\textbf{Spearman's Rank Correlation (Sp. Corr).} To facilitate comparison with previous methods\cite{14}\cite{20}\cite{2}\cite{13}, we use Sp. Corr to evaluate our method. Sp. Corr is commonly used to measure the correlation between predicted scores and actual scores, ranging from -1 to 1. A higher score indicates stronger agreement between predicted rankings and ground-truth rankings. The Sp. Corr between two sequences $x$ and $y$ is calculated as:
\begin{equation}
\rho = \frac{\sum_i (x_i - \bar{x})(y_i - \bar{y})}{\sqrt{\sum_i (x_i - \bar{x})^2 \sum_i (y_i - \bar{y})^2}},
\label{eq:spearman_corr}
\end{equation}
where $x_i$ and $y_i$ denote the rankings of the $i$-th sample in the predicted and ground-truth sequences, respectively, and $\bar{x}$, $\bar{y}$ are their corresponding mean rankings. In our AQA experiments, $x$ represents the ranking of the predicted scores generated by the model for all samples, while $y$ refers to the ranking of the ground-truth scores given by the expert judges. Therefore, Sp. Corr. reflects the correlation between the ranking order of the model’s predicted action quality scores and that of the ground-truth expert scores. To report an overall performance score across multiple action types, we follow prior work\cite{14}\cite{20}\cite{13}, and compute the average Sp. Corr using Fisher’s z-transformation, which ensures statistical robustness when aggregating correlation coefficients across different distributions.

\subsection{Implementation Details}
As detailed in Section 3.1, we first divide video frames, optical flow, and audio into identical, non-overlapping segments, each containing 32 frames, with segments shorter than 32 frames padded with zeros. Then, we use three pre-trained backbone networks to extract features from each modality separately. To optimize feature extraction, we employed features from previous multimodal AQA work\cite{20}, which were fine-tuned to adapt to the requirements of AQA. Finally, based on these extracted features, we conducted overall training of the multimodal local query encoder module and the two-level evaluation module. In addition, to ensure stable training, similar to \cite{14}\cite{20}\cite{13}, we normalize the labels to keep them within the range [0, 1].

In the experimental setup, each modality-specific branch comprises two stacked decoders, with each decoder having eight attention heads and two layers. The dropout rate is set to 0.1 for the RG dataset and 0.2 for the Fis-V dataset, respectively. The number of queries is fixed at 5 across all datasets, the output dimension is set to 512, and the learnable parameter $\tau$ is initialized to 0.07. We use the AdamW optimizer to train the model, with learning rates configured as 9e-4 for the RG dataset and 9e-5 for the Fis-V dataset. A cosine learning rate decay strategy is applied, and the batch size is set to 32. Our code is implemented in PyTorch and all experiments are conducted on a single NVIDIA A100 GPU.

\setlength{\tabcolsep}{5.535pt}
\renewcommand{\arraystretch}{1.3}
\begin{table}[!t]
\caption{The Sp. Corrs of our method are compared with other existing unimodal AQA methods on the RG and Fis-V datasets. Fisher's z-value is used to compute the average Sp. Corr across actions, where a higher value indicates better performance. The best results are highlighted in bold.}
\label{tab:1}
\renewcommand{\arraystretch}{1.3}
\setlength{\tabcolsep}{4.5pt}
\resizebox{\textwidth}{!}{
\begin{tabular}{|c|c|c|c|c|c|c|c|c|c|}
\hline
\multirow{2}{*}{Method} & \multirow{2}{*}{Features} & \multicolumn{5}{c|}{Rhythmic Gymnastics} & \multicolumn{3}{c|}{Fis-V} \\
\cline{3-10}
 & & Ball & Clubs & Hoop & Ribbon & \textbf{Avg.} & TES & PCS & \textbf{Avg.} \\
\hline
C3D+SVR~\cite{2} & C3D~\cite{35} & 0.357 & 0.551 & 0.495 & 0.516 & 0.483 & 0.400 & 0.590 & 0.501 \\
\hline
MS-LSTM~\cite{11} & VST~\cite{36} & 0.621 & 0.661 & 0.670 & 0.695 & 0.663 & 0.660 & 0.809 & 0.744 \\
\hline
ACTION-NET~\cite{13} & VST~\cite{36} + ResNet~\cite{37} & 0.684 & 0.737 & 0.733 & 0.754 & 0.722 & 0.694 & 0.809 & 0.744 \\
\hline
GDLT~\cite{14} & VST~\cite{36} & 0.746 & 0.802 & 0.763 & 0.841 & 0.765 & 0.685 & 0.820 & 0.761 \\
\hline
LMAC-Net(Ours) & VST~\cite{36} + I3D~\cite{39} + AST~\cite{38} & \textbf{0.803} & \textbf{0.806} & \textbf{0.856} & \textbf{0.881} & \textbf{0.840} & \textbf{0.811} & \textbf{0.881} & \textbf{0.850} \\
\hline
\end{tabular}
}
\end{table}

\setlength{\tabcolsep}{5.535pt}
\renewcommand{\arraystretch}{1.3}
\begin{table}[!t]
\caption{The Sp. Corrs of our method are compared with other existing state-of-the-art multimodal methods on the RG and Fis-V datasets. Fisher's z-value is used to compute the average Sp. Corr across actions, where a higher value indicates better performance. The best results are highlighted in bold.}
\label{tab:2}
\renewcommand{\arraystretch}{1.3}
\setlength{\tabcolsep}{4.5pt}
\resizebox{\textwidth}{!}{
\begin{tabular}{|c|c|c|c|c|c|c|c|c|c|}
\hline
\multirow{2}{*}{Method} & \multirow{2}{*}{Features} & \multicolumn{5}{c|}{Rhythmic Gymnastics} & \multicolumn{3}{c|}{Fis-V} \\
\cline{3-10}
 & & Ball & Clubs & Hoop & Ribbon & \textbf{Avg.} & TES & PCS & \textbf{Avg.} \\
\hline
Joint-VA~\cite{40} & VST~\cite{36} + AST~\cite{38} & 0.719 & 0.674 & 0.749 & 0.820 & 0.746 & 0.751 & 0.844 & 0.802 \\
\hline
MSAF~\cite{41} & VST~\cite{36} + I3D~\cite{39} + AST~\cite{38} & 0.743 & 0.795 & 0.734 & 0.836 & 0.781 & 0.751 & 0.843 & 0.802 \\
\hline
UMT~\cite{42} & VST~\cite{36} + AST~\cite{38} & 0.725 & 0.588 & 0.678 & 0.823 & 0.714 & 0.716 & 0.822 & 0.774 \\
\hline
PAMFN~\cite{20} & VST~\cite{36} + I3D~\cite{39} + AST~\cite{38} & 0.757 & 0.825 & 0.836 & 0.846 & 0.819 & 0.754 & 0.872 & 0.822 \\
\hline
LMAC-Net(Ours) & VST~\cite{36} + I3D~\cite{39} + AST~\cite{38} & \textbf{0.803} & 0.806 & \textbf{0.856} & \textbf{0.881} & \textbf{0.840} & \textbf{0.811} & \textbf{0.881} & \textbf{0.850} \\
\hline
\end{tabular}
}
\end{table}

\setlength{\tabcolsep}{5.535pt}
\renewcommand{\arraystretch}{1.3}
\begin{table}[!t]
\footnotesize
\caption{Computational efficiency comparison with existing mainstream long-term AQA methods.}
\label{tab:Efficiency}
\renewcommand{\arraystretch}{1.3}
\setlength{\tabcolsep}{6pt}
\centering
\begin{tabular}{|c|c|c|c|}
\hline
Method & Params & FLOPs & Inference Time \\
\hline
ACTION-NET~\cite{13} & 3.54M & 0.227G & 1ms \\
GDLT~\cite{14} & 1.85M & 0.075G & 3ms \\
PAMFN~\cite{20} & 17.96M & 0.721G & 36ms \\
LMAC-Net(Ours) & 8.95M & 0.419G & 4ms \\
\hline
\end{tabular}
\end{table}

\subsection{Comparison with the State-of-the-art}
To demonstrate the effectiveness of our method, we compared it with existing unimodal and multimodal methods on the RG and Fis-V datasets, with results shown in Table \ref{tab:1} and Table \ref{tab:2}. Notably, among the multimodal methods, only PAMFN\cite{20} was specifically designed for AQA task and tested on these two datasets. To comprehensively evaluate the performance of our method in multimodal tasks, we also selected the best-performing multimodal methods from other tasks and conducted comparisons based on re-implemented results on these two datasets. All compared methods were evaluated under identical experimental settings (data processing, dataset splits, and evaluation metrics) as our method, with their implementation results directly cited from prior work\cite{20}.

\noindent\textbf{Comparison with unimodal AQA methods.} Table \ref{tab:1} shows a comparison of the Sp. Corr between our method and other unimodal AQA methods on the RG and Fis-V datasets. On the RG dataset, our method achieves higher Sp. Corr across all categories (Ball, Clubs, Hoop, Ribbon), with an average of 0.840, significantly outperforming GDLT's 0.765. On the Fis-V dataset, our method also achieves the highest average correlation of 0.850, which is noticeably better than GDLT's 0.820. Compared to other unimodal AQA methods, our approach shows significant advantages in evaluation accuracy. This is achieved by leveraging multimodal features to capture the alignment between the performer's actions and the rhythm in long-duration sports with background music. 

\noindent\textbf{Comparison with multimodal AQA methods.}
To validate the effectiveness and advancement of our proposed method in multimodal tasks, we compared it with the best-performing multimodal AQA method on the RG and Fis-V datasets, PAMFN~\cite{20}, as well as other advanced multimodal methods from different tasks (Joint-VA~\cite{40}, MSAF~\cite{41}, UMT~\cite{42}), as shown in Table \ref{tab:2}. Compared to PAMFN, our method achieves better performance on both the RG and Fis-V datasets with fewer parameters, reaching average Sp. Corr of 0.850 and 0.840, respectively. On the RG dataset, our performance remains competitive, and we significantly outperform PAMFN in the other three categories (Ball, Hoop, and Ribbon), while our correlation coefficient is slightly lower in the Clubs category. On the Fis-V dataset, our method achieves leading results in TES and PCS evaluations, with Sp. Corr of 0.811 and 0.881, respectively, indicating higher efficiency and accuracy in multimodal information fusion. Furthermore, the comparisons with other multimodal methods highlight the superiority and robustness of our approach. Our method consistently outperforms these alternatives across all evaluation metrics, demonstrating excellent performance in handling complex multimodal tasks. This success can be attributed to the design of a multi-layer decoder architecture, the effective fusion and alignment of different modalities based on multimodal temporal parsing.

\textbf{Efficiency Analysis.}
We re-implemented all mainstream long-term AQA models under the same experimental settings and evaluated their efficiency, with the results shown in Table \ref{tab:Efficiency}. Compared to other multimodal methods, our proposed method demonstrates significant advantages in computational efficiency, requiring only 0.419G FLOPs, 8.95M parameters, and 4ms inference latency, which are all substantially lower than those of PAMFN~\cite{20}. Notably, although our method incorporates more modalities and exhibits increased model complexity, its inference speed is nearly on par with some unimodal long-term AQA methods, and is sufficiently fast to fully satisfy real-world deployment requirements. This is mainly attributed to our efficient cross-modal feature fusion and alignment strategy, which maintains low computational overhead while significantly improving performance.

\subsection{Ablation Study}
In this section, we conduct ablation studies to validate the effectiveness of the multimodal local query encoder and the two-level score evaluation module in our method.

\setlength{\tabcolsep}{5.535pt} 
\renewcommand{\arraystretch}{1.3} 

\begin{table}[!t]
\footnotesize 
\caption{Ablation study on adding the multimodal local query encoder Module (MLQE) and different internal losses to the baseline model on the RG and Fis-V datasets. Fisher's z-value is used to compute the average Sp. Corr across actions, where a higher value indicates better performance.}
\resizebox{\textwidth}{!}{  
\begin{tabular}{|l|c|c|c|c|c|c|c|c|c|c|c|c|}
\hline
\multirow{2}{*}{Method} & \multirow{2}{*}{MLQE} & \multirow{2}{*}{$L_{\text{rank}}$} & \multirow{2}{*}{$L_{\text{sparsity}}$} & \multirow{2}{*}{$L_{\text{consistency}}$} & \multicolumn{5}{c|}{Rhythmic Gymnastics} & \multicolumn{3}{c|}{Fis-V} \\
\cline{6-10} \cline{11-13}
& & & & & Ball & Clubs & Hoop & Ribbon & \textbf{Avg.} & TES & PCS & \textbf{Avg.} \\
\hline
Baseline & $\times$ & $\times$ & $\times$ & $\times$ & 0.667 & 0.586 & 0.698 & 0.735 & 0.676 & 0.664 & 0.787 & 0.731 \\
& \checkmark & $\times$ & $\times$ & $\times$ & 0.695 & 0.649 & 0.723 & 0.823 & 0.730 & 0.686 & 0.824 & 0.765 \\
& \checkmark & \checkmark & $\times$ & $\times$ & 0.708 & 0.656 & 0.748 & 0.796 & 0.731 & 0.716 & 0.830 & 0.779 \\
& \checkmark & \checkmark & \checkmark & $\times$ & 0.711 & 0.654 & 0.755 & 0.806 & 0.735 & 0.720 & 0.826 & 0.779 \\
Ours & \checkmark & \checkmark & \checkmark & \checkmark & \textbf{0.789} & \textbf{0.731} & \textbf{0.832} & \textbf{0.827} & \textbf{0.797} & \textbf{0.744} & \textbf{0.857} & \textbf{0.808} \\
\hline
\end{tabular}
}
\label{tab:3}
\end{table}

\noindent\textbf{Ablation study of the multimodal local query encoder module.} In this work, we propose a multimodal local query encoder module (MLQE) based on RGB, optical flow, and audio modalities. By introducing multimodal ranking loss and multimodal sparsity loss, we constrain the internal learning of modality-specific decoders, thereby enhancing multimodal feature representation and capturing fine-grained temporal features. Additionally, we utilize multimodal consistency loss to constrain the attention center distance of different modality queries, based on temporal parsing queries, which promotes cross-modal information interaction and facilitates feature fusion and alignment.

To verify the effectiveness of the MLQE component design, we built a multimodal baseline network. This basic network concatenates multimodal features directly and performs regression scoring through a linear layer. We then incrementally incorporate various architecture components into this baseline network to assess their impact on model performance. The experimental results are shown in Table \ref{tab:3}.

After incorporating the MLQE component into the baseline, significant improvements were observed in both category-specific performance and overall average performance, demonstrating the effectiveness of the MLQE in feature extraction and fusion. Introducing multimodal ranking loss and sparsity loss further improved the overall performance, demonstrating that these losses effectively enhance the model's ability to distinguish features from key temporal segments. Additionally, after integrating the proposed multimodal consistency loss, the model's average performance on the RG dataset increased from 0.735 to 0.797, and on the Fis-V dataset, it improved to 0.808. These results emphasize the importance of maintaining consistency across multimodal features. Our method better aligns different modalities on key segments, thereby enhancing the model's performance in long-term AQA and improving its ability to process multimodal information effectively.

\setlength{\tabcolsep}{5.535pt} 
\renewcommand{\arraystretch}{1.3} 

\begin{table}[!t]
\footnotesize 
\caption{Exploration of final integration method for multimodal information outputted by specific modal branches in the multimodal local query encoder module on the RG and Fis-V datasets. Fisher's z-value is used to compute the average Sp. Corr across actions, where a higher value indicates better performance.}
\resizebox{\textwidth}{!}{  
\begin{tabular}{|c|c|c|c|c|c|c|c|c|}
\hline
\multirow{2}{*}{Method} & \multicolumn{5}{c|}{Rhythmic Gymnastics} & \multicolumn{3}{c|}{Fis-V} \\
\cline{2-6} \cline{7-9}
& Ball & Clubs & Hoop & Ribbon & \textbf{Avg.} & TES & PCS & \textbf{Avg.} \\
\hline
Cross-modal Collaborative Attention $\rightarrow$  \text{Concatenation}
& 0.764 & 0.684 & 0.757 & 0.804 & 0.756 & 0.724 & 0.844 & 0.792  \\

Cross-modal Collaborative Attention $\rightarrow$  \text{Summation} & 0.779 & 0.684 & 0.755 & 0.754 & 0.745 & 0.719 & 0.836 & 0.783 \\

Cross-modal Collaborative Attention $\rightarrow$  \text{Dynamic Weighting} & 0.786 & \textbf{0.742} & 0.823 & 0.817 & 0.794 & 0.734 & 0.818 & 0.779 \\

Concatenation $\rightarrow$  \text{Attention} & 0.752 & 0.695 & 0.800 & 0.803 & 0.766 & \textbf{0.761} & \textbf{0.871} & \textbf{0.823}  \\

Summation $\rightarrow$  \text{Attention} & \textbf{0.802} & 0.689 & 0.811 & 0.787 & 0.774 & 0.759 & 0.858 & 0.814  \\

Summation & 0.792 & 0.686 & 0.810 & 0.774 & 0.770 & 0.697 & 0.823 & 0.767 \\

Concatenation & 0.789 & 0.731 & \textbf{0.832} & \textbf{0.827} & \textbf{0.797} & 0.744 & 0.857 & 0.808 \\
\hline
\end{tabular}
}
\label{tab:4}
\end{table}

\noindent\textbf{Multimodal Feature Fusion Strategies.}
After processing each modality-specific branch and performing modality feature alignment, our method requires integrating the aligned query features from the modality-specific branches for score regression. To evaluate the effectiveness of different strategies for integrating modality information, we experimented with several common methods:
\begin{itemize}
    \item \textbf{Cross-modal Collaborative Attention}. The features from different branches serve as queries to interact with other modalities, extracting cross-modal correlated features through an attention mechanism. 
    \item \textbf{Concatenation}. The features from different branches are concatenated along the feature dimension to form a richer feature representation.
    \item \textbf{Summation}. The features from different branches are added element-wise to obtain the final feature representation.
    \item \textbf{Dynamic Weighting}. Summing input features weighted by dynamically learned coefficients, typically generated through a linear layer. These coefficients adaptively capture the importance of each feature. 
    \item \textbf{Attention}. The features from different branches are first concatenated and then modeled using a multi-head self-attention mechanism to capture temporal dependencies both within and across modalities. Finally, the outputs from all queries are aggregated through average pooling.
\end{itemize}

The final experimental results are shown in Table \ref{tab:4}. Although various integration strategies perform differently across categories, simple concatenation proved to be the most stable and balanced approach overall. It consistently achieved near-optimal or second-best results in most categories on both the RG and Fis-V datasets. While cross-modal collaborative attention combined with dynamic weighting offered slight advantages in certain categories, it significantly increased model complexity. The methods combining feature concatenation or summation with attention, although achieving the best and second-best results on the Fis-V dataset, showed relatively average performance on the RG dataset, failing to meet the expected results. This indicates that incorporating self-attention can be beneficial for further integrating multimodal features in specific cases, but shows highly dataset-dependent performance and significantly increases training time. In contrast, direct feature concatenation simplifies model complexity, achieving near-optimal performance with notably lower computational cost.

\begin{table}[!t]
\footnotesize 
\caption{The comparison of Sp. Corrs of our proposed model with different modality-specific branches enabled on the RG and Fis-V datasets.}
\resizebox{\textwidth}{!}{  
\begin{tabular}{|c|ccc|c|c|c|c|c|c|c|c|}
\hline
\multirow{2}{*}{Method} & \multicolumn{3}{c|}{Modality} & \multicolumn{5}{c|}{Rhythmic Gymnastics} & \multicolumn{3}{c|}{Fis-V} \\
\cline{2-9} \cline{10-12}
& RGB & Flow & Audio & Ball & Clubs & Hoop & Ribbon & \textbf{Avg.} & TES & PCS & \textbf{Avg.} \\
\hline
\multirow{3}{*}{unimodal} &\checkmark &  &  & 0.524 & 0.553 & 0.591 & 0.707 & 0.599 & 0.559 & 0.677 & 0.623 \\
&  & \checkmark & & 0.474 & 0.520 & 0.522 & 0.681 & 0.554 & 0.637 & 0.533 & 0.588  \\
&  &  & \checkmark & 0.245 & 0.273 & 0.416 & 0.325 & 0.315 & 0.525 & 0.561 & 0.543  \\
\hline
\multirow{4}{*}{mutimodal} & \checkmark & \checkmark & & 0.795 & 0.746 & 0.770 & 0.851 & 0.793 & 0.670 & 0.819 &  0.754  \\
& \checkmark & & \checkmark & 0.720 & 0.714 & 0.725 & 0.790 & 0.738 & 0.675 & 0.788 & 0.737 \\
& & \checkmark & \checkmark & 0.567 & 0.536 & 0.634 & 0.771 & 0.637 & 0.622 & 0.760 & 0.699  \\
& \checkmark & \checkmark & \checkmark & \textbf{0.803} & \textbf{0.806} & \textbf{0.856} & \textbf{0.881} & \textbf{0.840} &\textbf{0.811} & \textbf{0.881} & \textbf{0.850} \\
\hline
\end{tabular}
}
\label{tab:5}
\end{table}

\noindent\textbf{Multimodal Performance.} To validate the effectiveness of combining multimodal features for AQA tasks, we also analyzed the impact of incorporating features from different modalities on the model's performance, in order to explore the complementarity between different modalities. The experimental results are shown in Table \ref{tab:5}.

The results show that unimodal models generally perform poorly, especially when only using the audio modality. In this case, the model's average performance on the RG dataset is only 0.315, which almost fails to complete the AQA task. This indicates that the audio modality has significant limitations in capturing details and evaluating complex movements.

By progressively adding different modalities as supplements to the unimodal model, our model demonstrates significant improvements in Sp. Corr and average Sp. Corr performance across actions. Specifically, the combination of RGB and optical flow exhibits strong complementarity, enhancing the model's performance by approximately 0.1 to 0.2 compared to using RGB or optical flow individually and achieving results that are near optimal. However, the combination of RGB and audio has some limitations in capturing motion details, as audio fails to effectively complement the visual information, leading to significantly lower performance than that of the RGB and optical flow combination. Similarly, the combination of optical flow and audio suffers from the absence of key visual cues, resulting in the worst performance among the multimodal combinations.

Regardless of whether it's the RGB unimodal, optical flow unimodal, or their combinations, adding the audio modality significantly improves performance. Particularly, in the combination of RGB and optical flow, which are strong complementary modalities, the addition of the audio modality boosts the model's average performance by 0.047 and 0.096 on the RG and Fis-V datasets, respectively, reaching the best performance achieved so far. This demonstrates that while the audio modality cannot fully replace visual information, it is a powerful supplement that enhances the accuracy of AQA evaluations.

\begin{table}[!t]
\footnotesize 
\caption{Sp. Corr of One-Stage Score Evaluation and Two-Level Score Evaluation Strategies on the RG and Fis-V datasets.}
\resizebox{\textwidth}{!}{  
\begin{tabular}{|c|c|c|c|c|c|c|c|c|c|}
\hline
\multirow{2}{*}{} & \multirow{2}{*}{Method} & \multicolumn{5}{c|}{Rhythmic Gymnastics} & \multicolumn{3}{c|}{Fis-V} \\
\cline{3-10} 
&  & Ball & Clubs & Hoop & Ribbon & \textbf{Avg.} & TES & PCS & \textbf{Avg.} \\
\hline
\multirow{2}{*}{One-Stage Score Evaluation} & MLP & 0.783 & 0.694 & 0.758 & 0.799 & 0.761 & 0.710 & 0.845 & 0.787 \\
& Linear & 0.789 & 0.731 & 0.832 & 0.827 & 0.797 & 0.744 & 0.857 & 0.808 \\
\hline
\multirow{5}{*}{Two-Level Score Evaluation}& Average & 0.793 & 0.764 & 0.835 & 0.856 & 0.816 & 0.799 & 0.857 & 0.831  \\
& MLP & 0.796 & 0.773 & 0.725 & 0.789 & 0.771 & 0.780 & 0.843 & 0.814  \\
& Linear & 0.791 & 0.742 & 0.785 & 0.790 & 0.778 & 0.775 & 0.846 & 0.814 \\
& Attention & 0.739 & 0.676 & 0.623 & 0.695 & 0.683 & 0.644 & 0.796 & 0.728 \\
& Weight & \textbf{0.803} & \textbf{0.806} & \textbf{0.856} & \textbf{0.881} & \textbf{0.840} & \textbf{0.811} & \textbf{0.881} & \textbf{0.850} \\
\hline
\end{tabular}
}
\label{tab:6}
\end{table}

\noindent\textbf{Two-Level Score Evaluation.}
To validate the effectiveness of the proposed two-level score evaluation, we compared it with the commonly used one-stage score evaluation methods in AQA tasks and explored various score fusion techniques. The experimental results are shown in Table \ref{tab:6}.

In the one-stage regression approach, aggregated multimodal features are averaged along the query dimension before being directly regressed using either an MLP or a linear regression layer. Our experiments revealed that linear regression outperformed MLP. However, traditional regression methods require pooling before regressing the scores, which inevitably leads to the loss of feature information. 

To address this issue, our approach directly regresses the score for each query along the query dimension, thereby ensuring the complete representation of multimodal features. In the first stage of our approach, we chose linear regression, which performed better in one-stage regression, as the regression head for each query.

Based on the scores for each query $\{\hat{s}_1,\hat{s}_2,\ldots,\hat{s}_K\}$, we further fuse these scores. In the second stage of score fusion, we experimented with the following strategies:
\begin{itemize}
    \item \textbf{Simple Average (Average)}. The query scores are averaged to calculate the final score. 
    \item \textbf{Non-Linear Fusion (MLP)}. The query scores are transformed using a Multi-Layer Perceptron.
    \item \textbf{Linear Fusion (Linear)}. The query scores are transformed using a linear layer to generate the final score.
    \item \textbf{Attention Fusion (Attention)}. The query scores are processed using a self-attention mechanism to calculate their importance and are then weighted accordingly.
    \item \textbf{Adaptive Weight Fusion (Weight)}. The query scores are combined using a learnable weight vector $w\in\mathbb{R}^K$, and the weight for each query score is gradually optimized during training to obtain the final score through weighted fusion.
\end{itemize}

As shown in Table~\ref{tab:6}, among all score fusion strategies, the Adaptive Weight Fusion method achieved the best performance across all categories, and its average performance on both datasets outperformed other methods. Additionally, our proposed regression method outperformed traditional one-stage regression methods under all fusion strategies except for the Attention method. This indicates that our two-level score evaluation strategy, compared to most AQA methods that directly regress scores using global representations, is better at establishing a direct and reasonable mapping between features and the final score, thereby enhancing model performance.

\setlength{\tabcolsep}{5.535pt} 
\renewcommand{\arraystretch}{1.3} 

\begin{table}[!t]
\footnotesize 
\caption{Comparison of Sp. Corrs for different multimodal alignment strategies on the RG and Fis-V datasets.}
\resizebox{\textwidth}{!}{  
\begin{tabular}{|c|c|c|c|c|c|c|c|c|}
\hline
\multirow{2}{*}{Method} & \multicolumn{5}{c|}{Rhythmic Gymnastics} & \multicolumn{3}{c|}{Fis-V} \\
\cline{2-6} \cline{7-9}
& Ball & Clubs & Hoop & Ribbon & \textbf{Avg.} & TES & PCS & \textbf{Avg.} \\
\hline 
CA & 0.657 & 0.572 & 0.663 & 0.731 & 0.654 & 0.569 & 0.783 & 0.670 \\

DCA~\cite{praveen2024cross} & 0.696 & 0.679 & 0.623 & 0.752 & 0.686 & 0.687 & 0.799 & 0.743 
\\

MTC Loss~\cite{sun2022long}
& 0.734 & 0.701 & 0.818 & 0.762 & 0.754 & 0.709 & 0.811 & 0.760  \\

Ours & \textbf{0.803} & \textbf{0.806} & \textbf{0.856} & \textbf{0.881} & \textbf{0.840} & \textbf{0.811} & \textbf{0.881} & \textbf{0.850}  \\
\hline
\end{tabular}
}
\label{Alignment Strategies}
\end{table}

\noindent\textbf{Multimodal Alignment Strategies.}
To further highlight the advantages of our attention-based multimodal temporal alignment approach in multimodal tasks, we selected several representative alignment strategies from the field of multimodal learning for ablation experiments. As introduced in Section 2.3, these methods primarily focus on the alignment capability of multimodal features at key temporal points. Specifically, the strategies include:
\begin{itemize}
    \item \textbf{Cross-Attention (CA)}. Utilizes a bidirectional cross-modal attention mechanism, where features from each modality serve as queries and keys/values for one another, enabling soft and adaptive alignment across modalities through the attention mechanism.
    \item \textbf{Dynamic Cross-Attention (DCA)}. Builds upon traditional cross-modal attention alignment by introducing a dynamic gating mechanism, allowing the alignment process to adaptively adjust the contributions of the original and aligned features based on modality complementarity, thus enhancing the effectiveness of cross-modal alignment.
    \item \textbf{Multimodal Temporal Contrastive Loss (MTC Loss)}. Applies an InfoNCE-style contrastive loss across modalities at each temporal segment (query), explicitly aligning the multimodal feature representations at the same time step.   
\end{itemize}
As shown in Table \ref{Alignment Strategies}, our proposed alignment method consistently outperforms other mainstream approaches on both the RG and Fis-V datasets. Specifically, while CA can enhance information exchange across modalities, it lacks fine-grained temporal constraints and thus fails to achieve precise synchronization at key moments of action, resulting in the lowest overall performance. DCA improves the flexibility of alignment through dynamic gating and partially optimizes modality complementarity, but its ability to synchronize information over long-term actions remains limited due to the modeling capacity of the interaction layers. Notably, although MTC Loss and other CLIP-style contrastive losses perform well in multimodal retrieval scenarios, they essentially only achieve global representation alignment and overlook the temporal synchronization of action segments, leading to inferior results in AQA tasks compared to our method.

In contrast, our consistency loss explicitly constrains the attention centers of different modalities at key segments, thereby achieving action-level temporal alignment. Compared to simply pulling feature similarity closer in the embedding space, this provides more targeted guidance and more effectively promotes the coordinated focus of multimodal information at critical moments.

\subsection{Qualitative Analysis}    
\noindent\textbf{Attention-Driven Multimodal Alignment.} We visually demonstrate the significant role of the proposed attention-driven multimodal alignment method in multimodal learning. In Figure 5, we track and present the dynamic changes in the attention centers of different modalities during the training process for three randomly selected training samples from the TES and PCS evaluation tasks in the Fis-V dataset. Without applying our proposed multimodal consistency loss (Figures 5(a)-(b)), which is the foundation of the multimodal alignment method, the attention centers of the three modalities exhibit obvious spatial and temporal dispersion and disorder. This not only reduces the efficiency of information interaction between modalities but also potentially leads to confusion during the fusion process. In contrast, after introducing our multimodal consistency method (Figures 5(c)-(d)), the attention centers of different modalities gradually become highly consistent.

More importantly, relying on the model's temporal parsing capability, we observe that as training progresses, the attention centers of queries from different modalities tend to align sequentially along the temporal axis. This demonstrates not only the method's ability to achieve efficient information interaction across modalities but also its effectiveness in capturing key temporal segments in long-term complex actions, significantly enhancing the model's understanding and fusion of long-term temporal features.

\begin{figure}[H]
    \centering
    \begin{subfigure}[b]{0.95\textwidth} 
        \centering
\includegraphics[width=\textwidth]{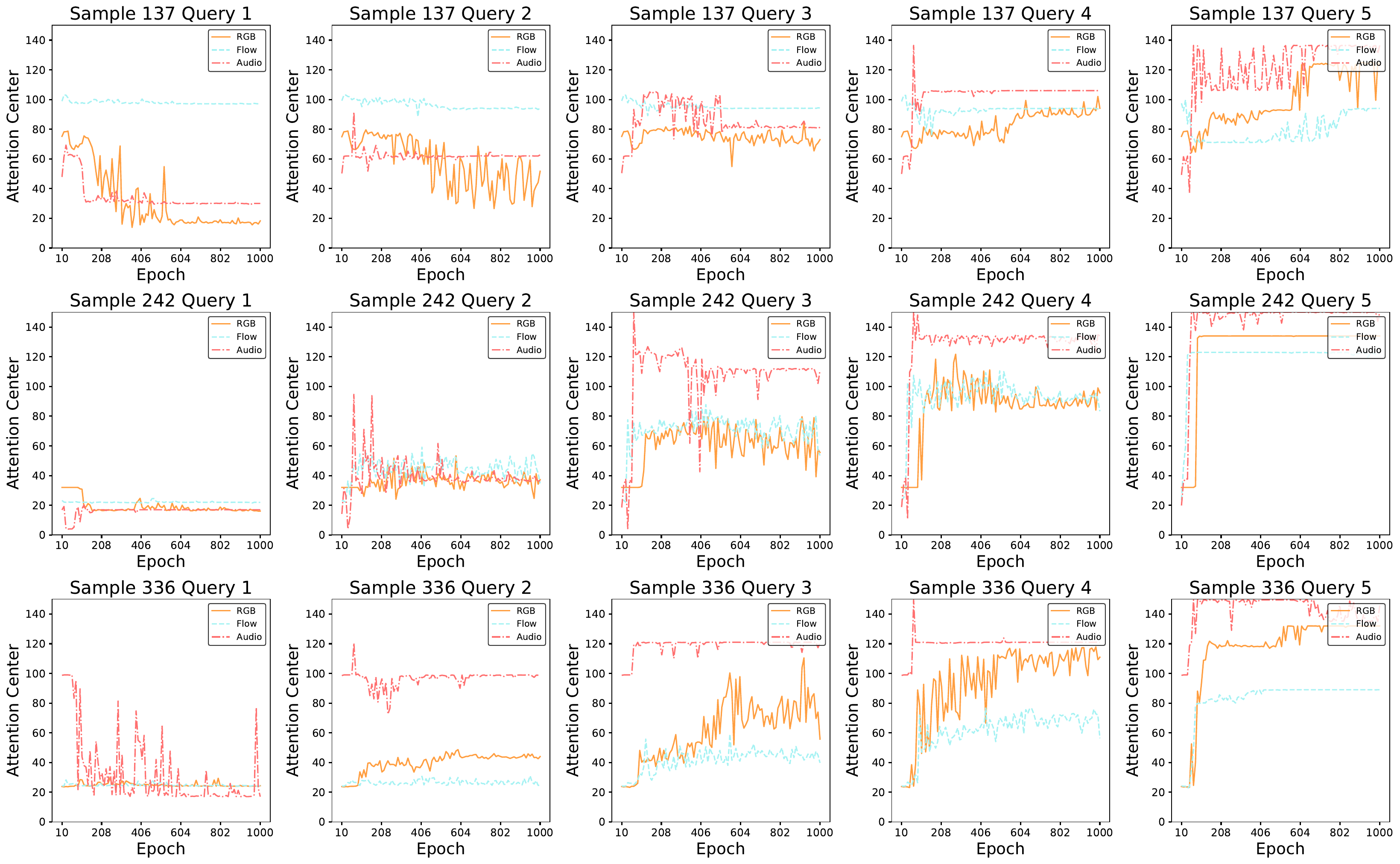} 
\captionsetup{labelformat=empty} 
        \caption{(a) TES}
    \end{subfigure}
    
    \begin{subfigure}[b]{0.95\textwidth} 
        \centering        \includegraphics[width=\textwidth]{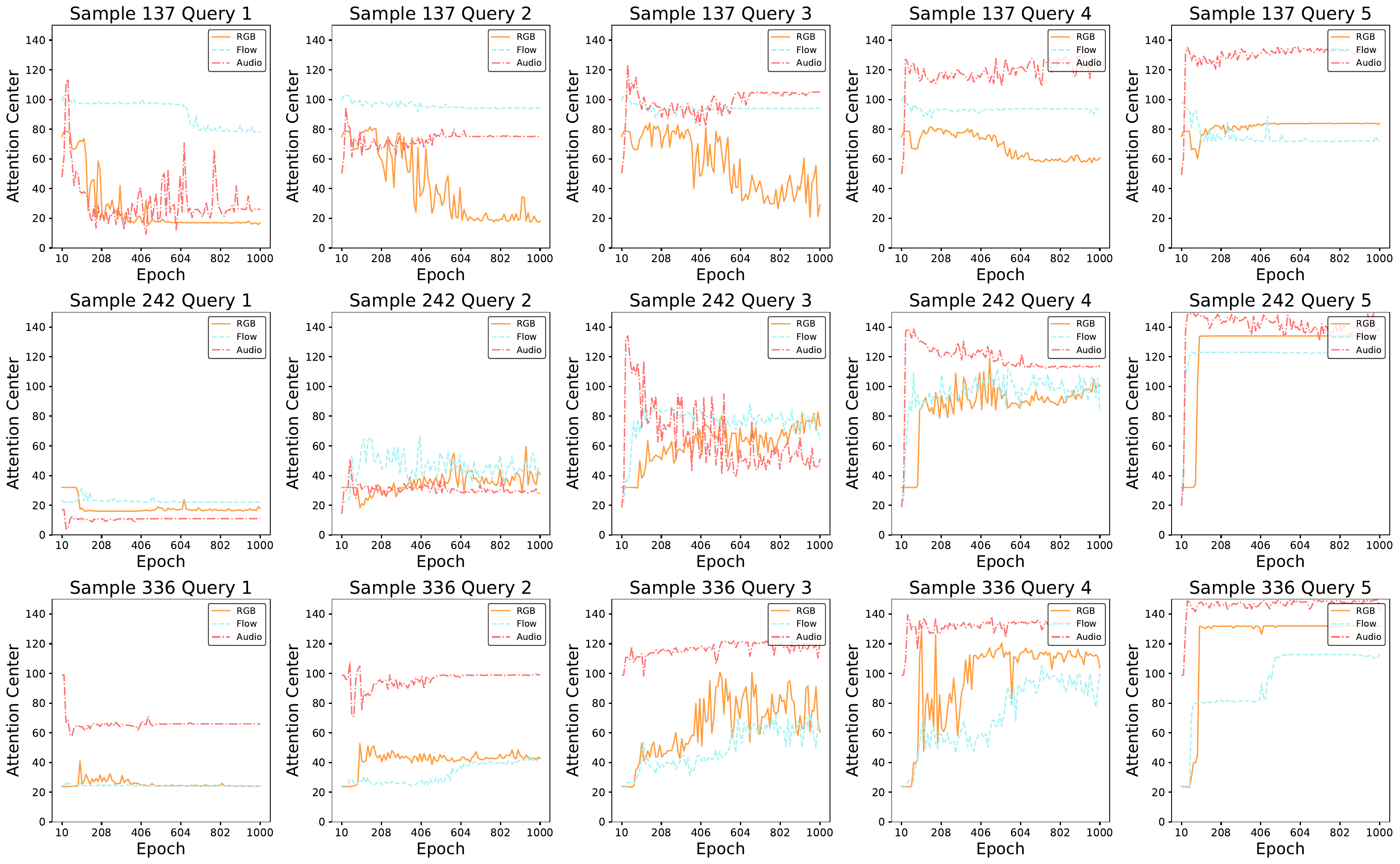}
\captionsetup{labelformat=empty} 
        \caption{(b) PCS}
    \end{subfigure}
\end{figure}
\newpage
\captionsetup[figure]{labelformat=empty}
\begin{figure}[H]
\centering
\begin{subfigure}[b]{0.95\textwidth} 
        \centering
\includegraphics[width=\textwidth]{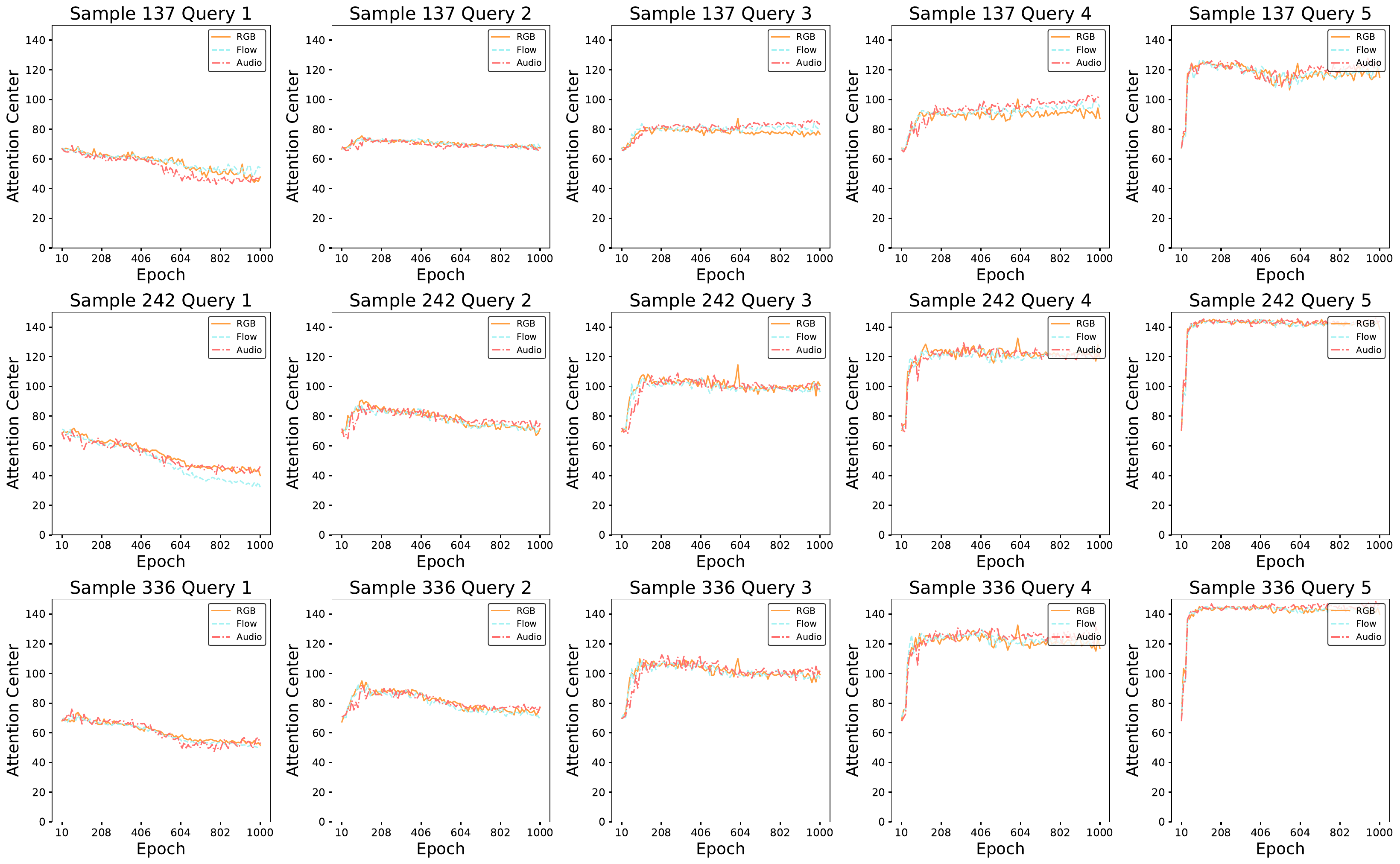} 
\captionsetup{labelformat=empty} 
        \caption{(c) TES-Multimodal Alignment}
    \end{subfigure}
    \begin{subfigure}[b]{0.95\textwidth} 
        \centering\includegraphics[width=\textwidth]{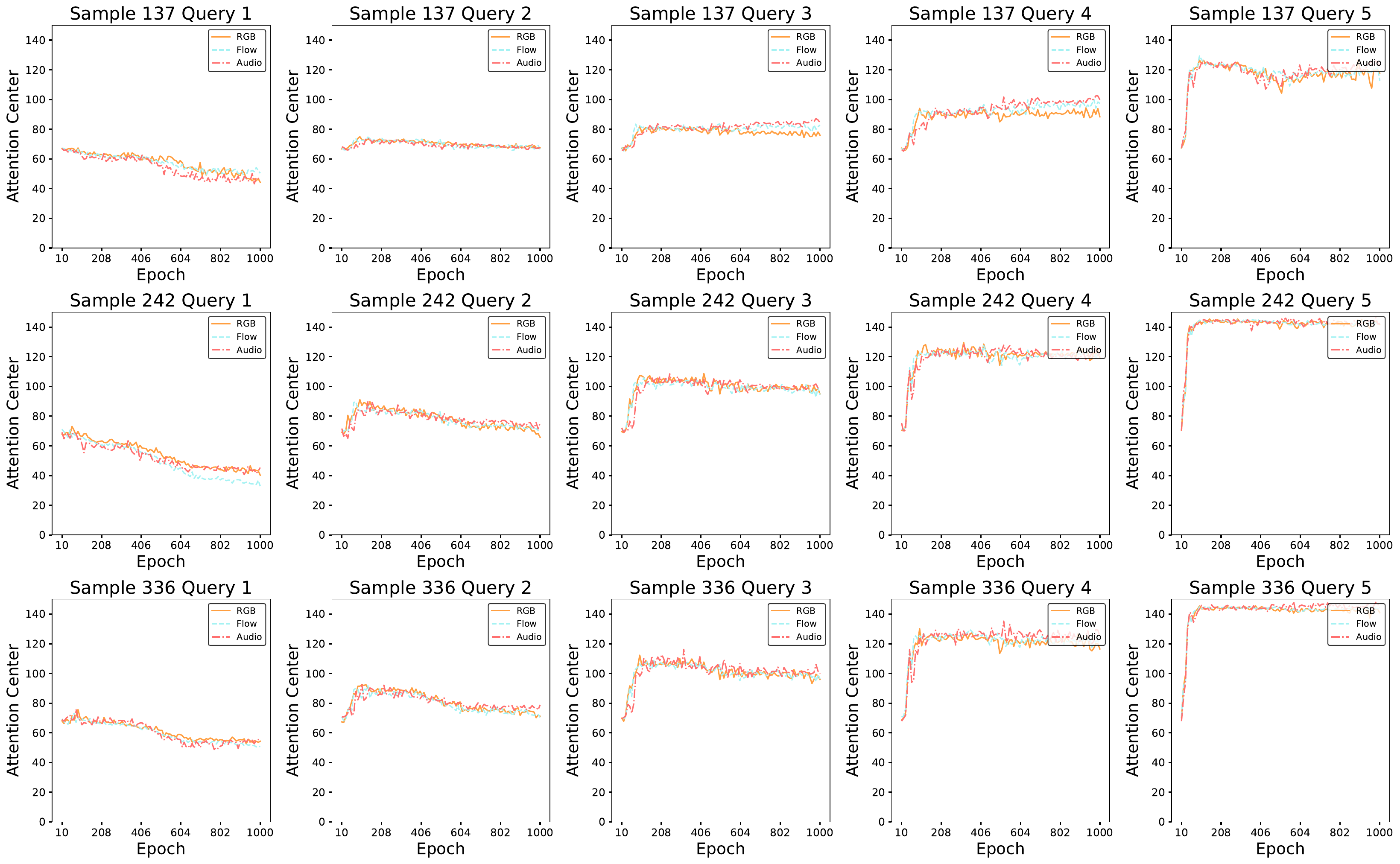} 
\captionsetup{labelformat=empty} 
        \caption{(d) PCS-Multimodal Alignment}
    \end{subfigure}
    \caption{Figure 5: Changes in attention centers of modality-specific queries across three randomly selected training samples from Fis-V dataset during the training process. (a)-(b) show the results without multimodal alignment, while (c)-(d) illustrate the effects after introducing our multimodal alignment method. }
    \label{fig:Figure_5}
\end{figure}
To further quantify the alignment effect, we used the average cosine similarity between modalities as a measurement metric for the tracked training samples and presented the final results in Figure 6. The X-axis represents training epochs, the Y-axis indicates sample indices, and the color intensity corresponds to the average cosine similarity value. The cosine similarity between RGB-Flow, RGB-Audio, and Flow-Audio is calculated, and their mean is taken as the overall alignment metric. Higher cosine similarity indicates greater convergence of feature distributions across modalities, reflecting better coordination of attention-concentrated regions. As shown in Figure 6, without the alignment method (Figures 6(a)-(b)), the cosine similarity between different modalities remains consistently low and fluctuates significantly during training. This suggests that even with temporal parsing of individual modality features, the advantages of multimodal learning are difficult to fully realize without constraining cross-modal consistency. Conversely, with our multimodal alignment method, the cosine similarity between different modalities approaches its maximum value during training, indicating that the distribution of attention centers across modalities on the temporal axis gradually converges and achieves high consistency.
\begin{figure}[t]
    \centering
    \begin{subfigure}[t]{0.495\textwidth} 
        \centering
        \includegraphics[width=\textwidth]{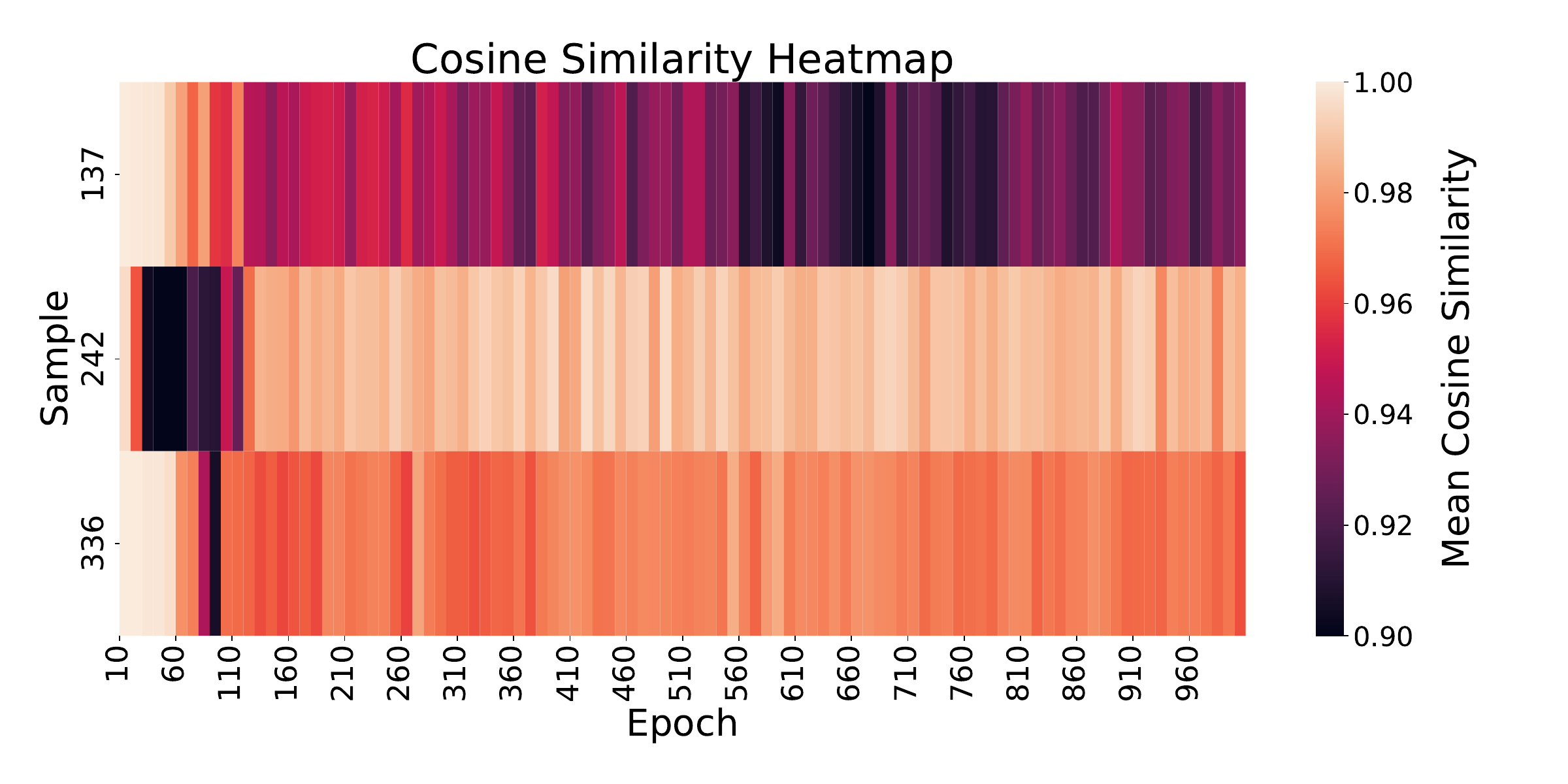} 
        \caption{TES}
    \end{subfigure}
    \hfill
    \begin{subfigure}[t]{0.495\textwidth} 
        \centering
        \includegraphics[width=\textwidth]{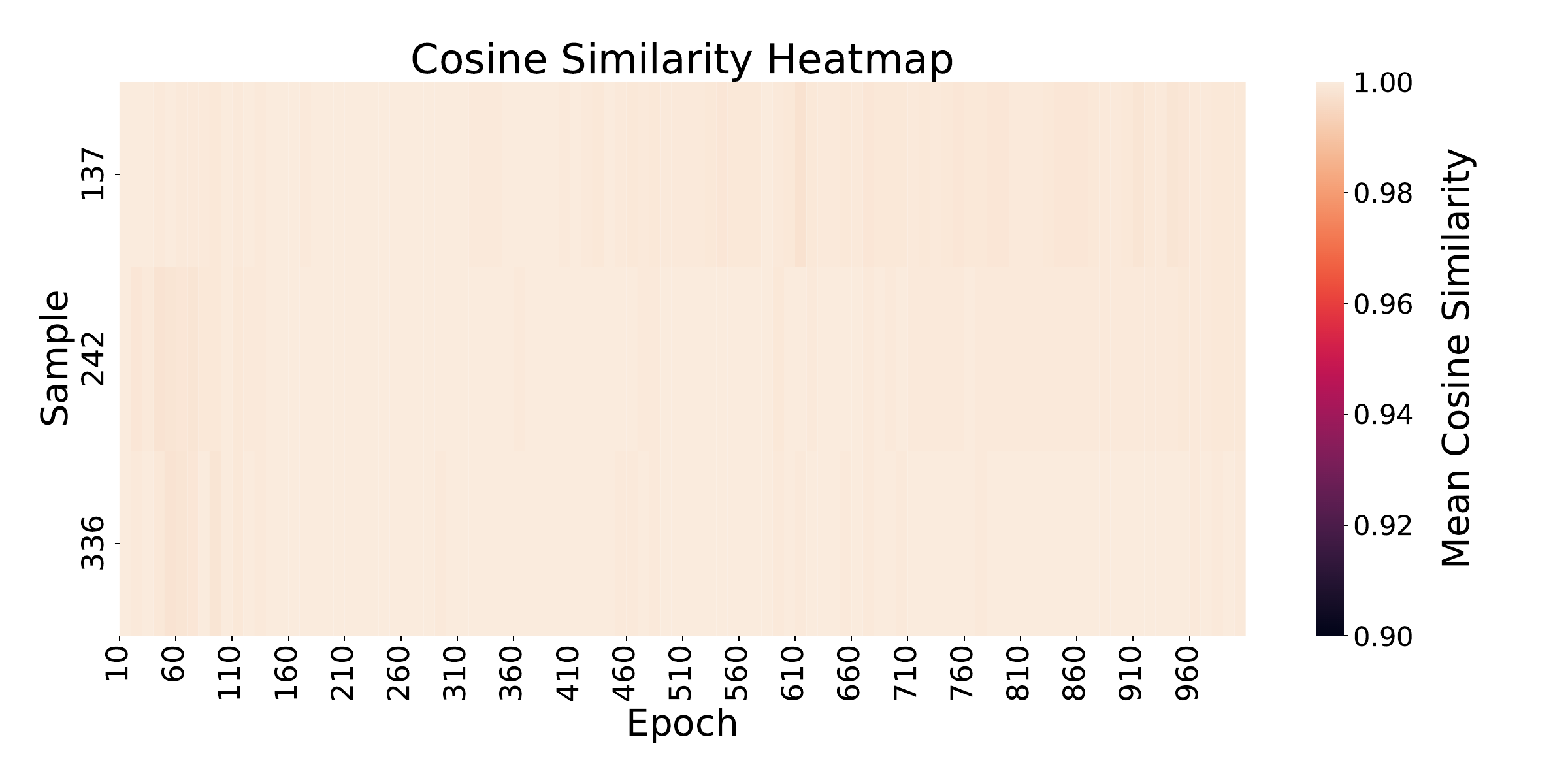}
        \caption{TES-Multimodal Alignment}
    \end{subfigure}
    
    \hfill

    \begin{subfigure}[t]{0.495\textwidth} 
        \centering
        \includegraphics[width=\textwidth]{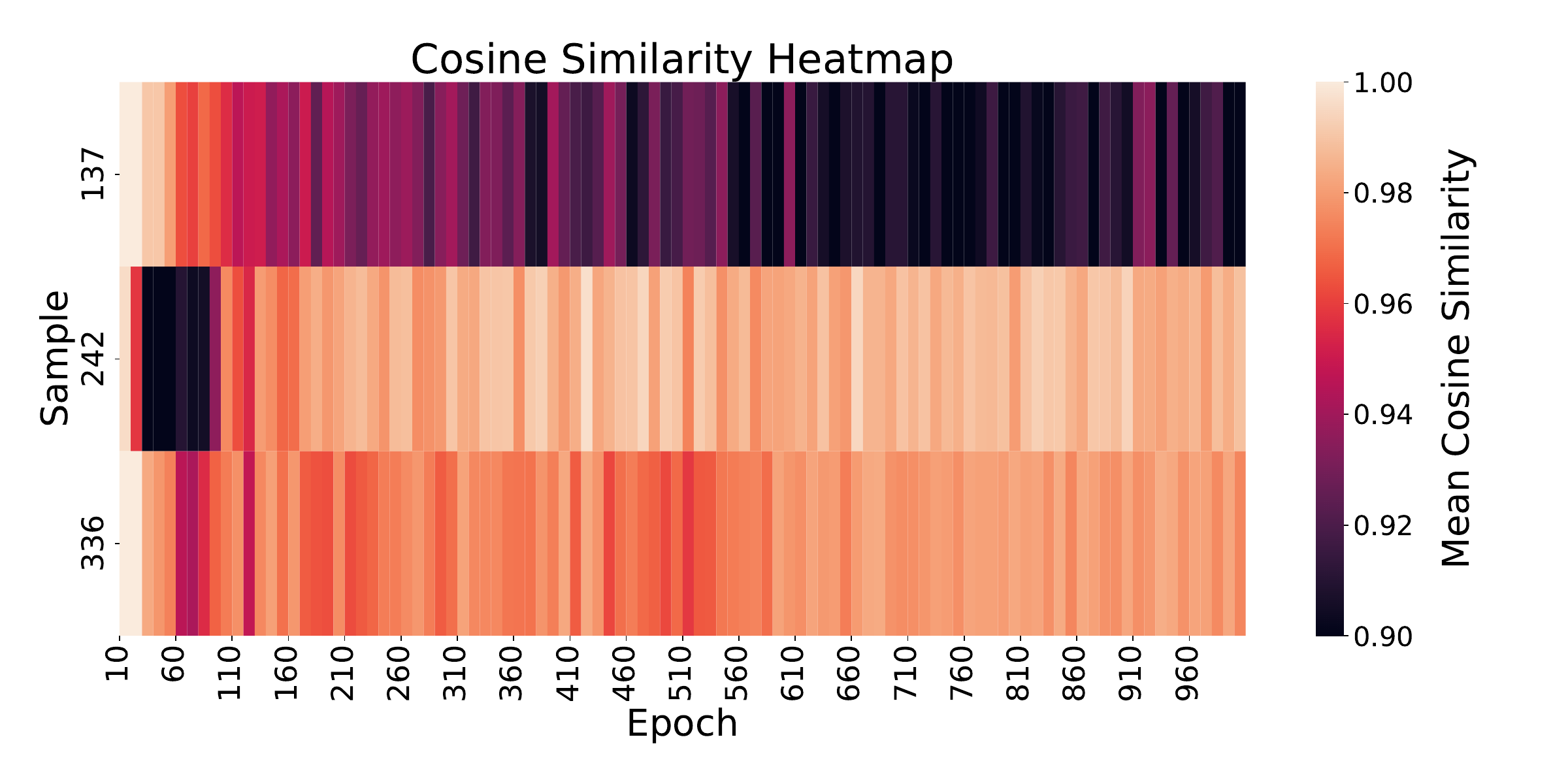} 
        \caption{PCS}
    \end{subfigure}
    \hfill
    \begin{subfigure}[t]{0.495\textwidth} 
        \centering
        \includegraphics[width=\textwidth]{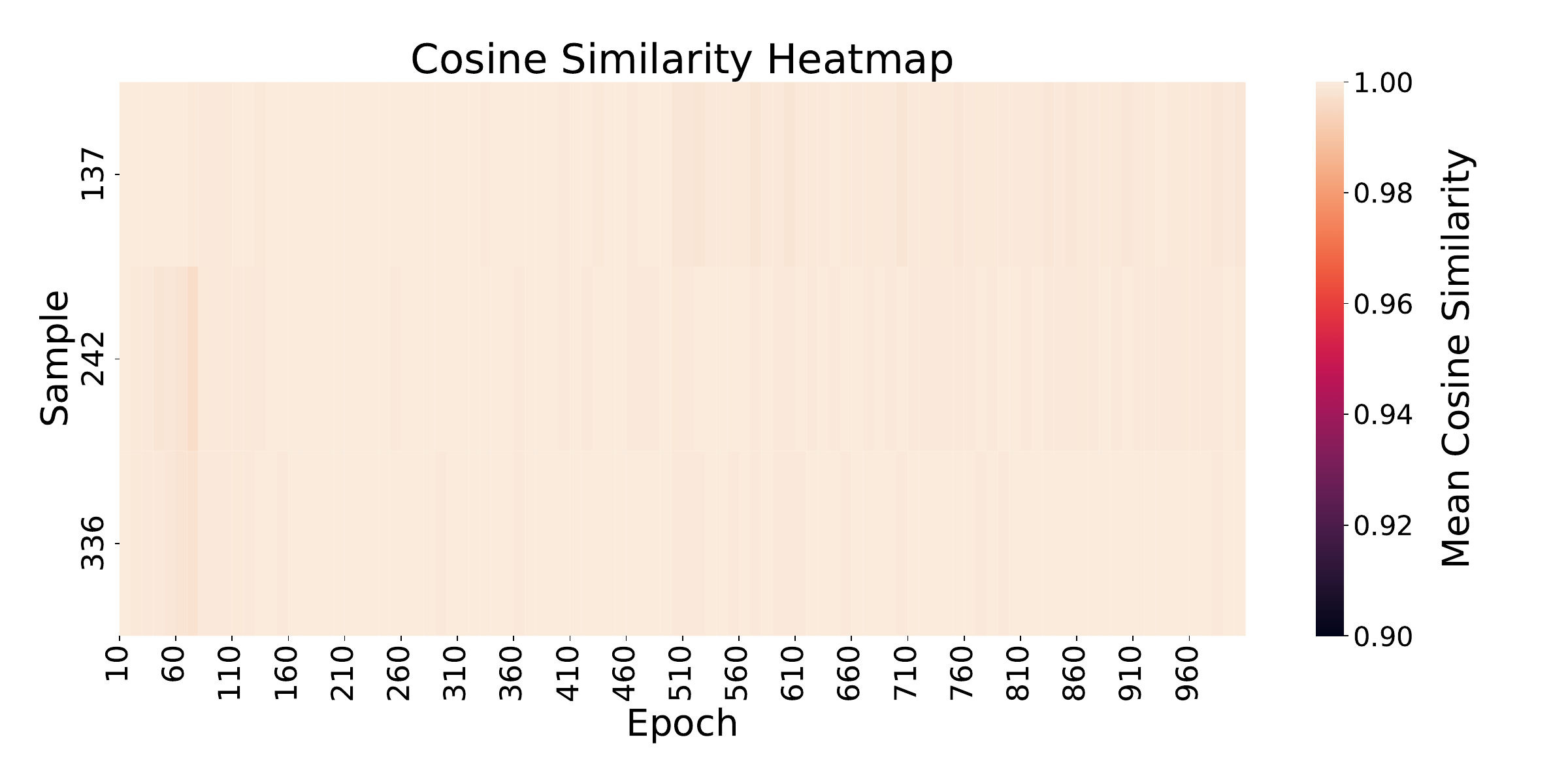} 
        \caption{PCS-Multimodal Alignment}
    \end{subfigure}
    
    \caption{Figure 6: Heatmaps comparing the three-modality alignment performance on the Fis-V dataset. (a)-(d) illustrate the trends in mean cosine similarity with and without the modality alignment method, where higher values indicate better alignment.}
    \label{fig:Figure_6}
\end{figure}

\begin{figure*}[t]
    \centering
    \begin{subfigure}[t]{0.495\textwidth} 
        \centering
        \includegraphics[width=\textwidth]{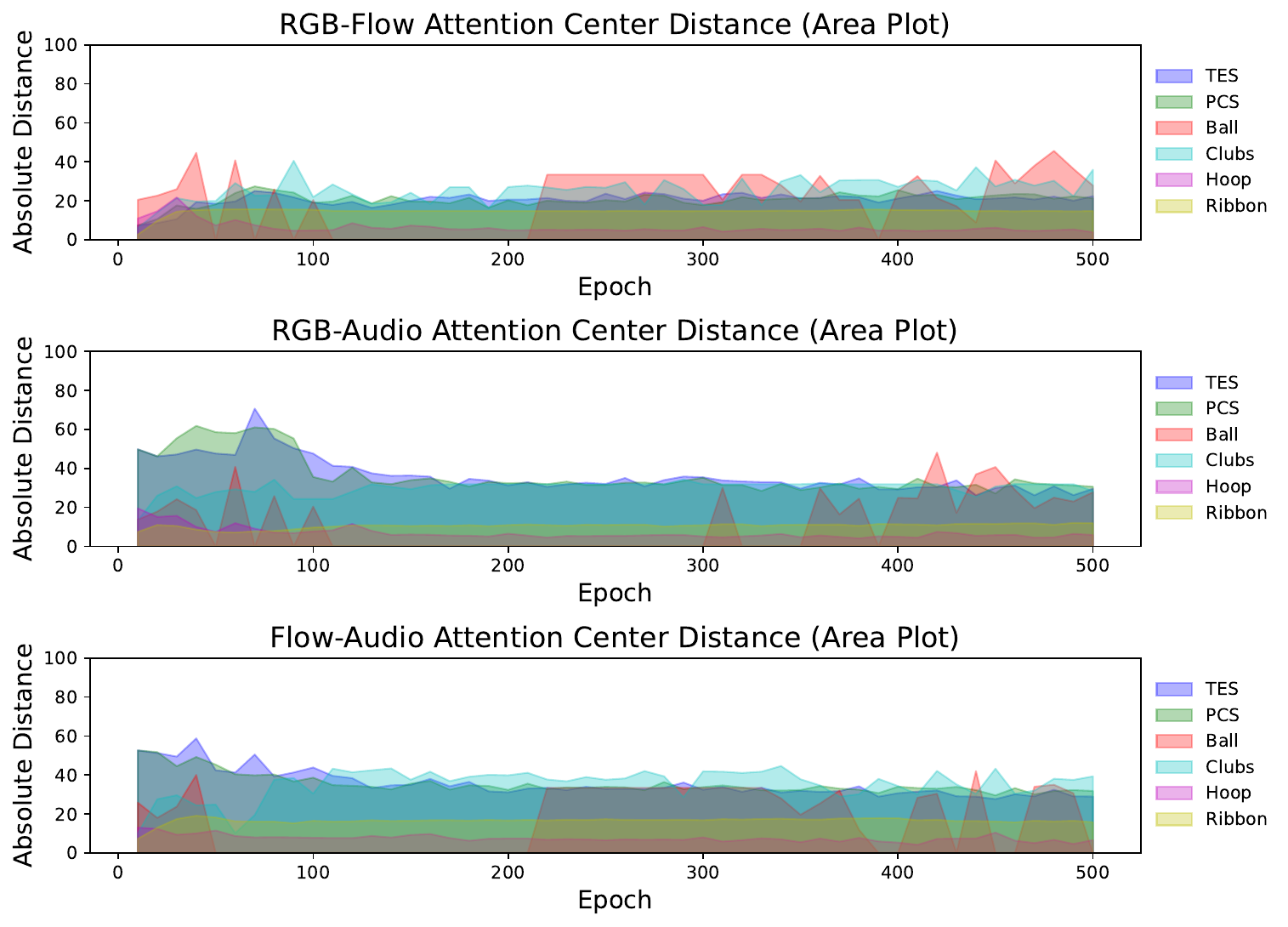} 
        \caption{Our method without modality alignment}
    \end{subfigure}
    \hfill
    \begin{subfigure}[t]{0.495\textwidth} 
        \centering
        \includegraphics[width=\textwidth]{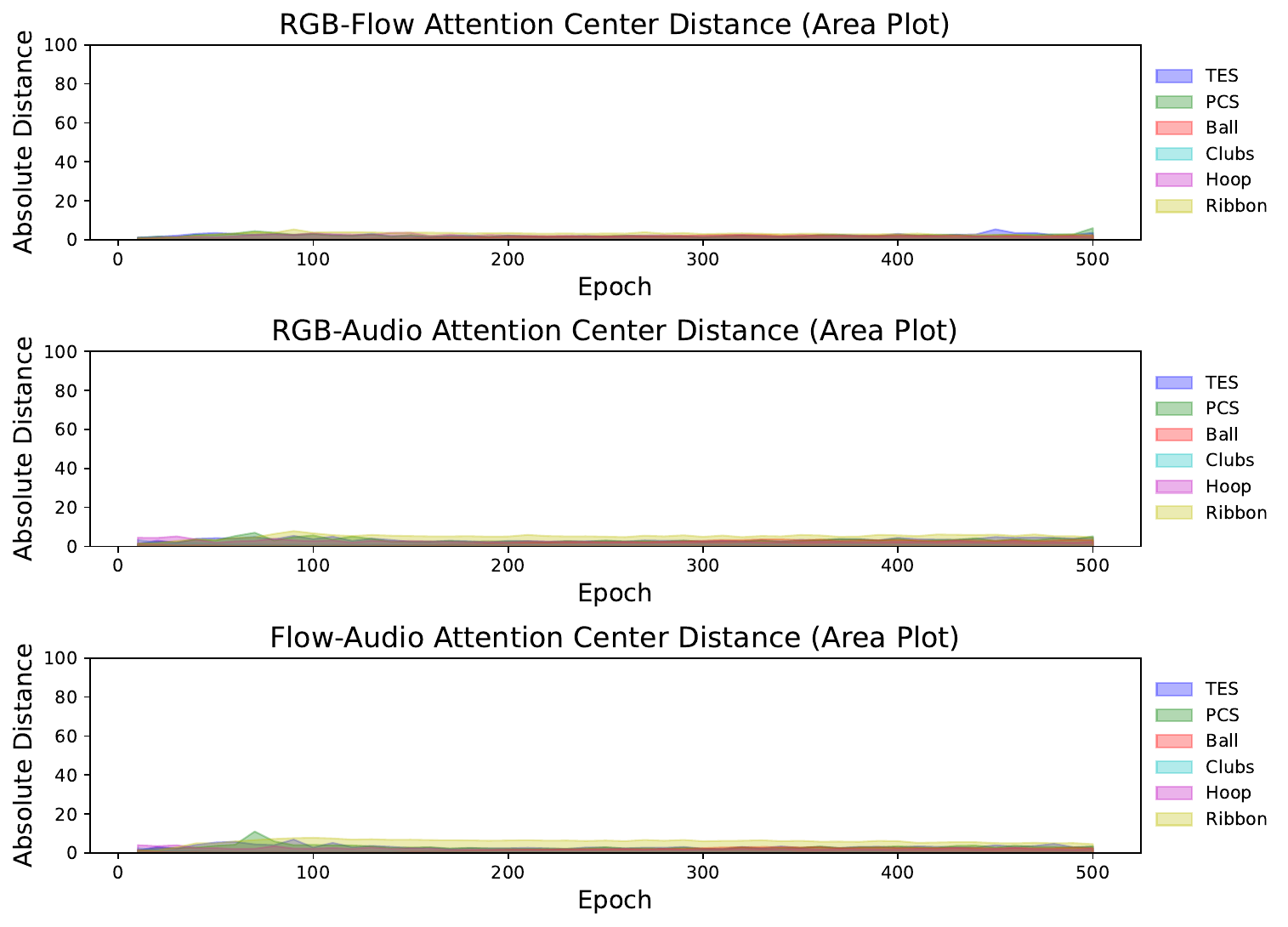} 
        \caption{Our method with modality alignment}
    \end{subfigure}
    
    \caption{Figure 7: The area charts depict the trends and magnitudes of the absolute distances between attention centers of different modalities over training epochs. Subfigure (a) shows results without the proposed attention-based modality alignment method, while subfigure (b) shows results with the method applied.}
    \label{fig:Figure_7}
\end{figure*}

\begin{figure}[h] 
    \centering
    \includegraphics[width=0.95\columnwidth]{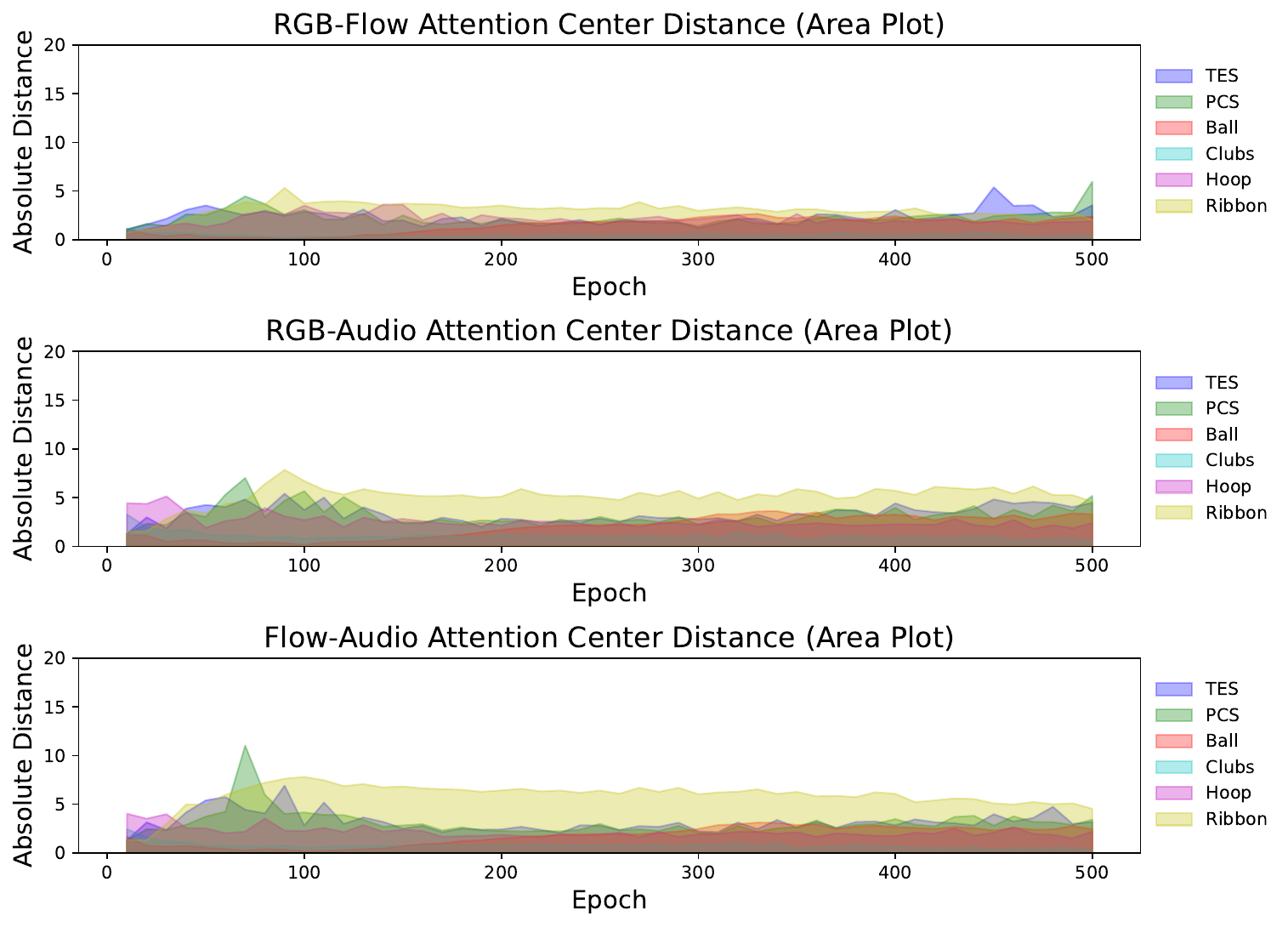} 
    \caption{Figure 8: The alignment accuracy between modalities measured by the attention center difference. The subplots illustrate the distance variations between RGB-Flow, RGB-Audio, and Flow-Audio modalities. }
    \label{fig:Figure_8}
\end{figure}

Furthermore, to provide a more intuitive demonstration of alignment precision, we present the corresponding visualization results in Figures 7 and 8. In this experiment, we selected three training samples, identical to those in Figures 5 and 6, from the entire dataset and tracked and calculated the distances between attention centers for different modality pairs (RGB-Flow, RGB-Audio, Flow-Audio). The average distance for the same modality pair across the three samples was calculated for each dataset to reflect the overall consistency level of different modality pairs during training.

In Figure 7, when alignment is not applied (Figure 7(a)), the distances between modalities fluctuate significantly throughout training. After introducing our alignment method (Figure 7(b)), the absolute distances between attention centers are significantly reduced. Figure 8 further highlights this alignment effect, showing high overall alignment precision for all modality pairs. Among them, the distance for the RGB-Flow modality pair is the smallest, indicating stronger complementarity and fusion characteristics between visual modalities. The distances for RGB-Audio and Flow-Audio are slightly larger but still achieve satisfactory alignment. This indicates that, although there are inherent differences between the audio and visual modalities, our alignment method seeks to reduce such differences by forcing the attention focus points of different modalities to converge as closely as possible, thereby significantly enhancing the degree of multimodal fusion.

\begin{figure*}[t]
    \centering    \includegraphics[width=\textwidth]{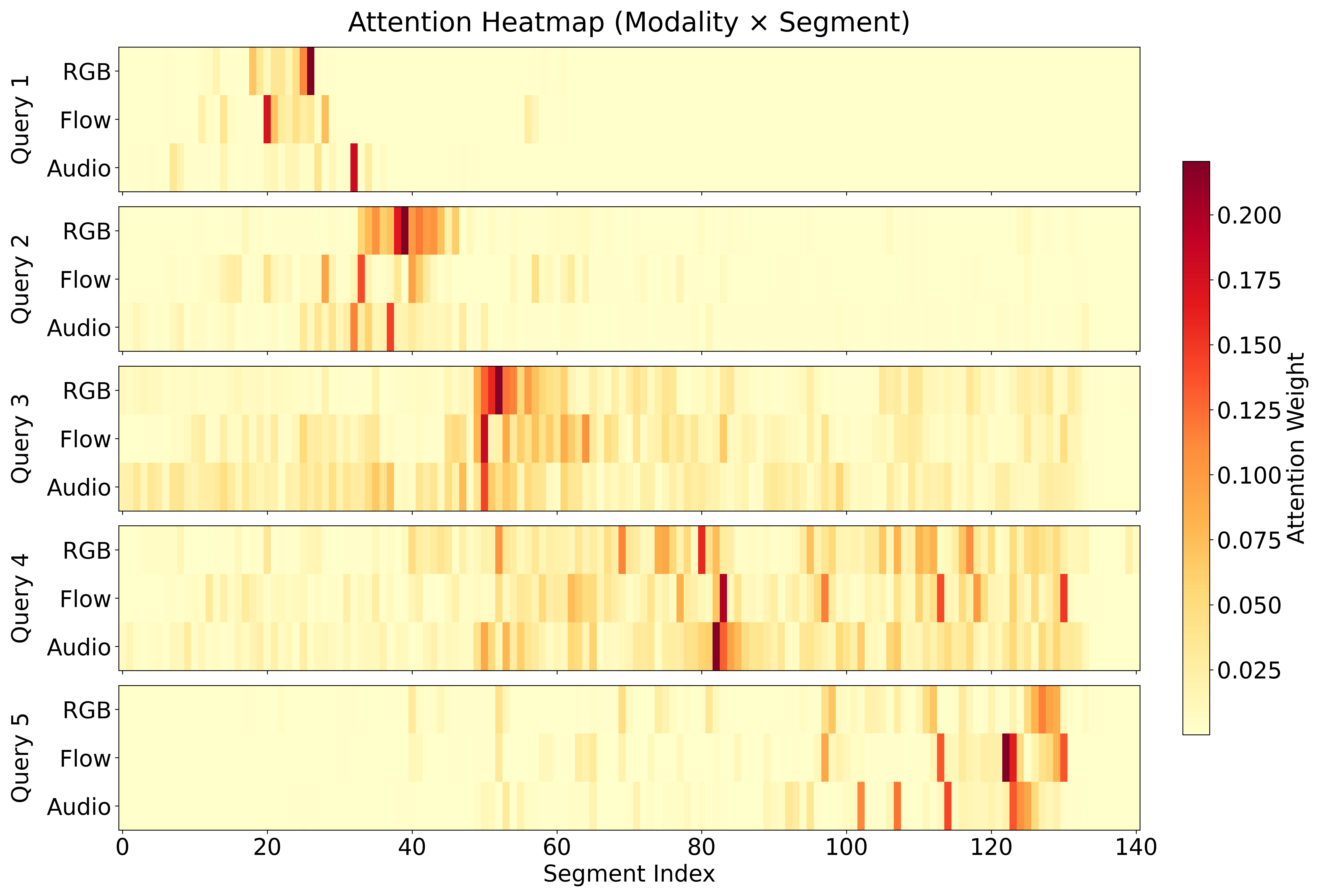} 
    \caption{Figure 9: Visualization of the cross-attention weights in the multimodal local query encoder module. Each major row represents the attention distribution of a query across different temporal segments, and the three sub-rows correspond to the RGB, Flow, and Audio modalities. The horizontal axis denotes the segment index, while the vertical axis indicates the input modality.}
    \label{fig:Figure_9}
\end{figure*}

\begin{figure*}[t]
    \centering    \includegraphics[width=\textwidth]{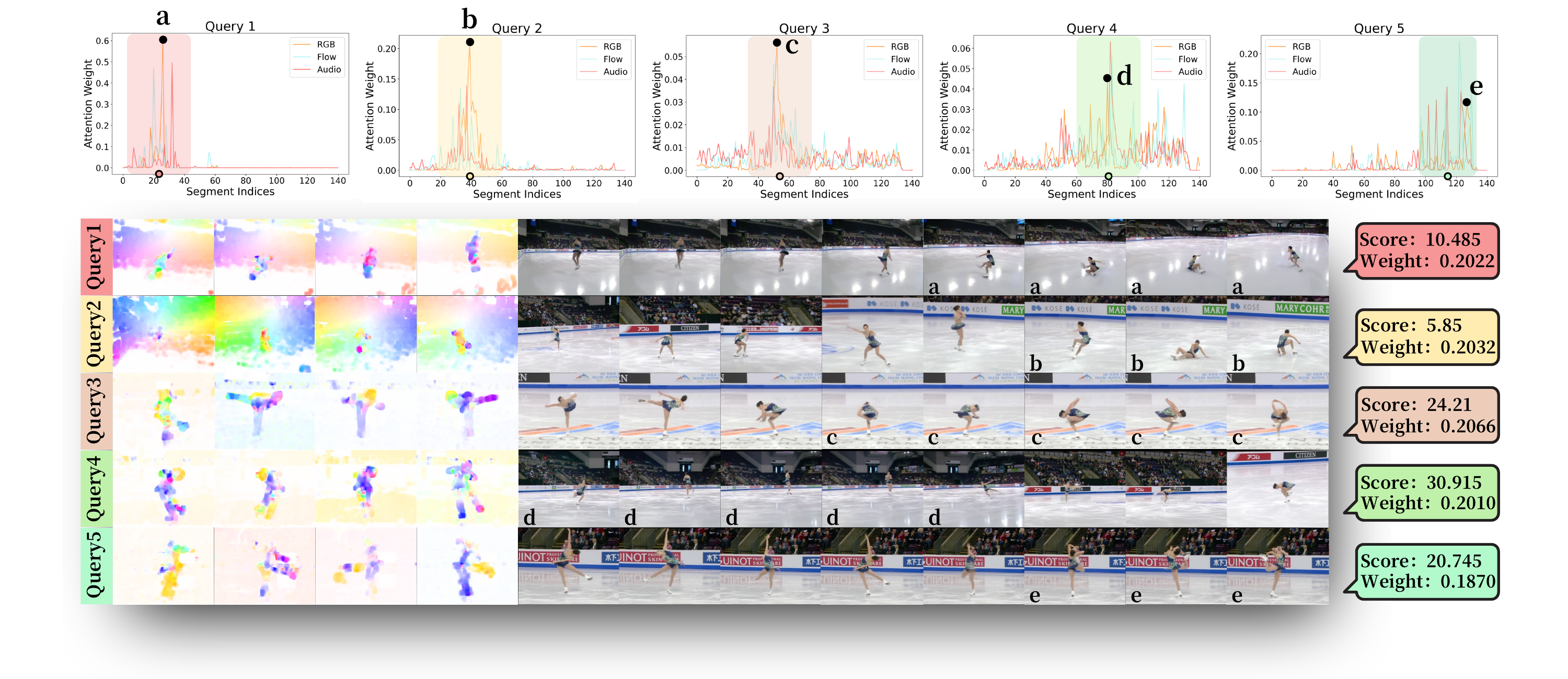} 
    \caption{Figure 10: Visualization of the cross-attention weights in the final decoder layer of each modality-specific branch within the multimodal local query encoder module, together with the corresponding key frames.}
    \label{fig:Figure_10}
\end{figure*}

\noindent\textbf{Visualization of Cross-attention Weights.} To examine whether each query accurately captures the temporal pose features across multiple modalities, we extracted the attention weights from the final decoder layer of each modality-specific branch within the multimodal local query encoder module. These weights reflect the degree of attention each query assigns to different temporal segments. Based on the 308th sample from the Fis-V dataset (TES), we provide relevant visualizations in Figures 9 and 10, including attention weight heatmaps and curves. Additionally, key frames were selected for visualization to help understand the model's ability to capture key action features in the temporal dimension and its cross-modal coordination.

In Figure 9, with the introduction of multimodal ranking loss and sparsity loss, we observe that each query exhibits distinct attention regions along the temporal axis, and the overall attention distribution is relatively sparse, only a few temporal segments receive high weights. Furthermore, thanks to the introduction of multimodal consistency loss, different modalities for the same query are generally focused on the same temporal region, achieving collaborative focus of multimodal information. It is worth noting that, despite the alignment constraints, each modality still emphasizes different details within the same region, demonstrating the complementarity among modalities.

Furthermore, to comprehensively analyze the specific performance of LMAC-Net in terms of multimodal complementarity and temporal alignment, we present in Figure 10 the cross-attention weight curves of the final decoder layer for each modality branch, as well as the corresponding video frames. The first row displays the attention weight curves for 5 queries across the three modalities. The subsequent rows show video segments associated with these 5 queries and their corresponding query scores. For analysis, we focus on the 5 moments with the highest attention weights in the RGB modality, marked as a, b, c, d, and e, indicating the corresponding video segments. As shown in Figure 10, the attention weight curves of all modalities clearly highlight the peak distributions of attention across corresponding temporal regions. Even in complex action sequences, each query focuses highly on specific temporal segments containing critical action features and automatically aligns these high-information-gain regions across the RGB, flow, and audio modalities. More importantly, the attention weights exhibit notable synchronization across modalities during peak intervals, indicating that the model achieves consistent temporal representations across modalities, enabling better fusion of multimodal information and enhancing its understanding of complex action sequences. Specifically:
\begin{itemize}
    \item \textbf{Query 1} focuses on the initial stage of the performance. The RGB modality assigns the highest weight to the moment when the performer falls (a), clearly capturing the poor posture, while the flow modality exhibits weaker attention due to the lack of prominent dynamic features at that moment. However, during the gliding and airborne rotation phases prior to the fall, the flow modality’s weight increases, accurately capturing changes in gliding speed and rotational dynamics. Simultaneously, the audio modality responds to the synchronization of gliding movements with the music rhythm.
    \item \textbf{Query 2} highlights the performer’s interaction with the music through elegant gliding movements, reflected in increased weights for both the flow and audio modalities, capturing smooth dynamic trajectories and strong alignment with the music rhythm. However, during a poorly executed landing after a jump (b), the RGB modality provides direct visual feedback.
    \item \textbf{Query 3} focuses on high-dynamic regions involving continuous rotations (c). All three modalities are assigned high weights: the flow modality captures the rotational trajectories and intensity, the RGB modality ensures pose continuity, and the audio modality responds to peaks in the music rhythm.
    \item \textbf{Query 4} covers the transitions between multiple actions. During the performer’s gliding dance, both flow and audio modalities increase in weight, capturing dynamic trajectories and music rhythm. Meanwhile, the RGB modality captures the aerial rotation and perfect landing (d) with precision. Subsequent rotational movements are primarily represented by the flow and audio modalities, reflecting the action’s dynamics and rhythm.
    \item \textbf{Query 5} focuses on the finale, characterized by various challenging rotational movements (e). The RGB modality accurately captures details and pose stabilization during high-difficulty rotational movements, while the flow and audio modalities synergistically respond to the dynamics and rhythm changes of the rotations.
\end{itemize}

\noindent\textbf{Score Interpretability.} 
To explore the interpretability of the scoring evaluation strategy, we extracted and visualized the scores and weights corresponding to each query from the two-level score evaluation module (as shown in Figure 9). These scores provide a detailed representation of the model's evaluation of action quality across different temporal segments.

As illustrated in Figure 9, the model assigns higher scores to temporal segments with better execution of high-difficulty actions (e.g., Query 3 and Query 4), demonstrating its ability to accurately identify and emphasize high-quality temporal segments. In contrast, for segments with poorer performance (e.g., Query 1 and Query 2), the model adopts a more balanced scoring strategy. For example, in Query 1, although the score drops due to a landing error, the earlier excellent gliding and rotation-jumping performance mitigates the negative impact of the error. In Query 2, however, the model assigns a significantly lower score due to multiple landing errors and overall poor action quality.

Regarding weight distribution, our model tends to allocate weights more evenly for scoring tasks like figure skating, while still adjusting them based on the action quality of the temporal segments. Notably, in Query 1, the model slightly reduces the weight to avoid disproportionate impacts on the overall score due to brief errors. In contrast, for Query 2, the model not only lowers the score but also increases the weight for this segment to highlight the adverse effects of consecutive errors on the overall action quality.

We believe that this flexible and balanced scoring mechanism provided by the two-level score evaluation offers greater interpretability and robustness in the model's decision-making process.

\section{Conclusion}
In this work, we proposed Long-term Multimodal Attention Consistency Network (LMAC-Net) to address the task of long-term action quality assessment (AQA). Specifically, we introduced query vectors into the RGB, flow, and audio modalities, using multi-layer transformer decoders to independently decode the temporal features of each modality. To achieve effective cross-modal alignments and capture complementary information, we applied attention-center distance constraints across modalities, ensuring that different modalities focus on similar key moments along the temporal dimension, thereby enhancing the consistency of multimodal learning. Additionally, we proposed a two-level score evaluation strategy: first assigning local scores to each query, then fusing them into final global scores. Experiments on two publicly available long-term AQA datasets demonstrate that our method effectively addresses numerous challenges in long-term multimodal AQA, including complex temporal dependencies, multimodal feature alignments and interactions, and fine-grained interpretability of evaluation results. We believe that this work provides insights not only for multimodal AQA but also for maintaining consistency across modalities in other multimodal learning tasks.

\section{Acknowledgments}
This work has been supported by Beijing National Science Foundation under Grant No. 9234029 and the Fundamental Research Funds for the Central Universities under Grant No. 2024JCYJ004.

\bibliographystyle{unsrt}
\bibliography{reference}

\end{document}